\documentclass[sigconf, screen]{acmart}
\AtBeginDocument{%
  }

\setcopyright{acmlicensed}
\copyrightyear{2018}
\acmYear{2018}
\acmDOI{XXXXXXX.XXXXXXX}
\acmConference[Conference acronym 'XX]{Make sure to enter the correct
  conference title from your rights confirmation email}{June 03--05,
  2018}{Woodstock, NY}
\acmISBN{978-1-4503-XXXX-X/2018/06}

\acmSubmissionID{2182}

\usepackage{enumitem}
\usepackage{pifont}
\usepackage{multirow}
\usepackage{subcaption}
\usepackage[table]{xcolor} 
\usepackage{booktabs}
\usepackage{graphicx}
\usepackage{makecell}
\usepackage{float}   
\usepackage{hyperref}
\usepackage{multirow}
\usepackage{amsmath}
\usepackage{caption}
\usepackage{algorithm}
\usepackage{algpseudocode}
\usepackage{bbm}

\begin{document}

\title{Dual-Branch Cross-Projection Debiasing through Diffusion-based Disentanglement}


\author{Xiangqian Zhao}
\affiliation{%
  \institution{Xidian University}
  \city{Shaanxi}
  \postcode{710071}
  \country{China}
}

\author{Xinyang Jiang}
\affiliation{%
  \institution{Microsoft Research Asia}
  \city{Shanghai}
  \postcode{200232}
  \country{China}
}

\author{Zhipeng Xu}
\affiliation{%
  \institution{Xidian University}
  \city{Shaanxi}
  \postcode{710071}
  \country{China}
}

\author{Lingfeng He}
\affiliation{%
  \institution{Xidian University}
  \city{Shaanxi}
  \postcode{710071}
  \country{China}
}

\author{Zilong Wang}
\affiliation{%
  \institution{Microsoft Research Asia}
  \city{Shanghai}
  \postcode{200232}
  \country{China}
}

\author{Dongsheng Li}
\affiliation{%
  \institution{Microsoft Research Asia}
  \city{Shanghai}
  \postcode{200232}
  \country{China}
}

\author{De Cheng}
\affiliation{%
  \institution{Xidian University}
  \city{Shaanxi}
  \postcode{710071}
  \country{China}
}

\author{Nannan Wang}
\affiliation{%
  \institution{Xidian University}
  \city{Shaanxi}
  \postcode{710071}
  \country{China}
}

\begin{abstract}
Foundation models trained on biased datasets often rely on spurious correlations between target labels and non‑causal attributes, resulting in poor generalization on minority groups. Bias mitigation remains challenging due to two fundamental issues. First, when group labels are unavailable, existing group‑unsupervised methods typically infer spurious attributes implicitly from model behavior, making it difficult to identify spurious factors that are semantically aligned with real‑world biases. 
Second, even with pseudo spurious supervision, most existing debiasing methods follows a single-branch design which operates within a single shared feature space, where target and spurious attributes are intrinsically entangled.
To address the first challenge, we introduce Confidence‑guided Bias Concept Mining (CBCM), which leverages diffusion‑disentangled, semantically grounded concept representations to identify reliable spurious attributes without attribute annotations.
To address the second challenge, we propose Dual‑branch Cross‑projection Debiasing (DCD), a prompt‑tuning framework that separates target and spurious representations into two branches and explicitly removes spurious information through cross null‑space projection while preserving target‑relevant semantics.
Extensive experiments on four benchmark datasets show that our method achieves state‑of‑the‑art worst group accuracy among group‑unsupervised approaches, while tuning at most 0.22\% of the model parameters. The source code is available in the supplementary materials.

\end{abstract}




\maketitle

\begin{figure}[t]
    \centering
    \includegraphics[width=0.48\textwidth]{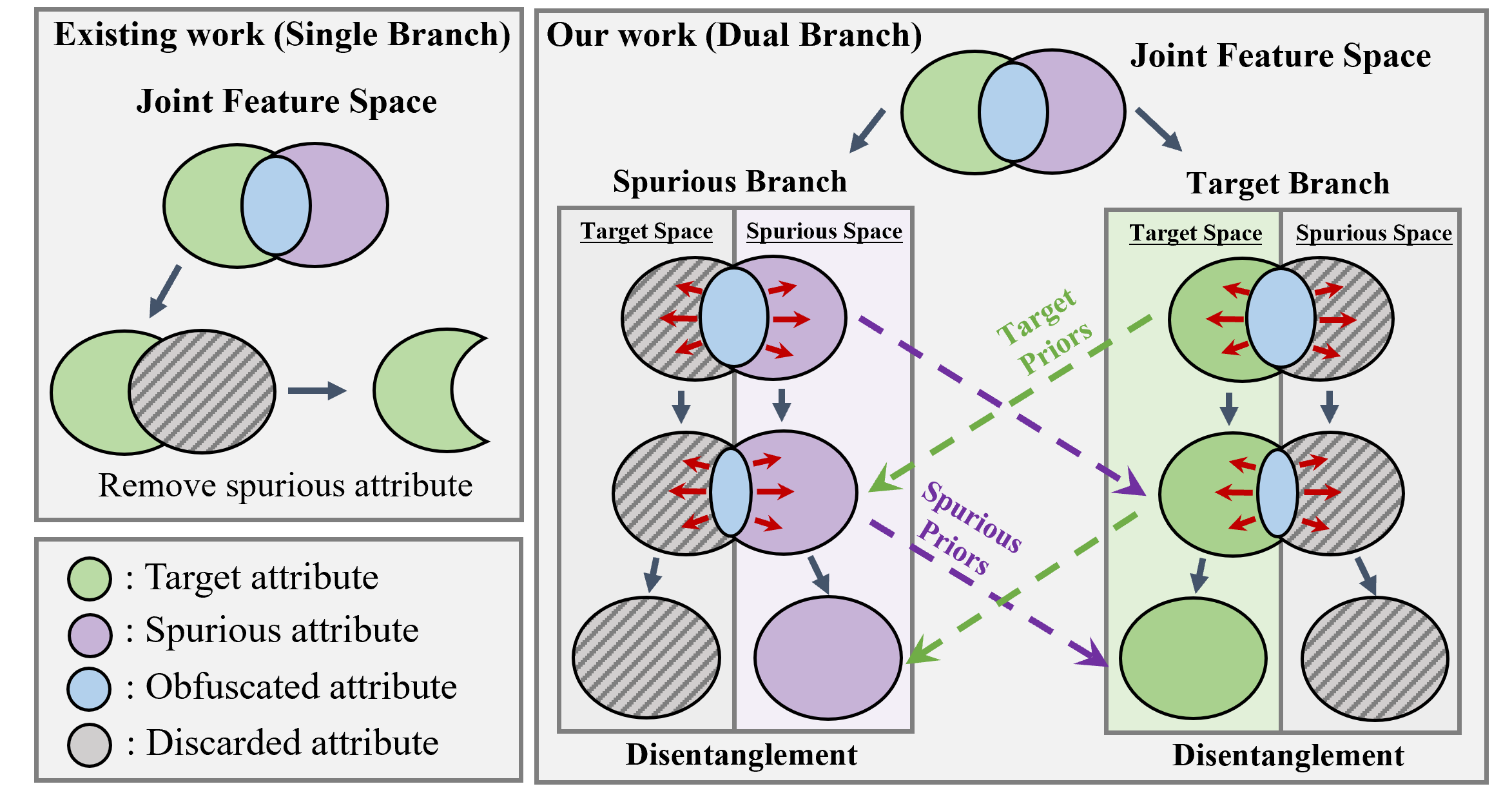}
    \caption{Unlike single-branch methods that may damage target information, our dual-branch design separates target and spurious factors and uses cross-projection to suppress spurious information while preserving target semantics.}
    \label{fig:intro}
\end{figure}

\section{Introduction}

In recent years, foundation models have been widely adopted across vision tasks. However, they are typically trained to optimize average performance over the entire training set. When the data are biased, models may inadvertently rely on spurious correlations between easy to learn, non-target attributes and target labels. Although such shortcuts can improve overall accuracy, they often lead to poor generalization on data where these spurious correlations are absent. As a result, models with strong average performance may still perform poorly on certain subsets of data. This challenge has motivated growing interest in fairness, particularly in settings that seek to maintain robust performance across different subpopulations.





A central challenge in bias mitigation is obtaining reliable spurious attribute supervision when group labels are unavailable. 
Most group-unsupervised methods tackle this problem by first training a biased model and then deriving pseudo spurious labels from its behavior, e.g., prediction confidence, training dynamics, or implicitly disentangled representations~\cite{liu2021justtraintwiceimproving,pezeshki2024discoveringenvironmentsxrm,zhang2024amend}. Although effective in practice, such supervision is often indirect and not semantically grounded, since it is inferred from model behavior or optimization-induced latent factors rather than explicit attributes. 
Consequently, it remains unclear whether the identified “spurious” representations correspond to real-world bias factors such as color, or gender, making it difficult to reliably remove the true source of bias.


Recent advances in diffusion models provide a new opportunity to overcome these limitations~\cite{Tao2024EncDiff}. 
Diffusion-based disentanglement is particularly effective because iterative denoising progressively isolates distinct factors of variation, while cross-attention grounding encourages each concept to align with semantically coherent visual attributes, thereby yielding semantically meaningful and disentangled concept representations.
Building on this property, we propose the first framework to exploit diffusion‑disentangled, semantically grounded concepts for debiasing. We introduce Confidence‑guided Bias Concept Mining (CBCM) to mine reliable pseudo spurious supervision without requiring attribute annotations. CBCM first trains a diffusion model equipped with a concept encoder, which decomposes each image into concept tokens, with each token capturing a distinct semantic attribute. We then train a biased classifier on these concept tokens, and spurious attributes are identified as the tokens that most strongly activate the classifier. These selected concepts are subsequently used as pseudo spurious labels to guide downstream debiasing.

Another key challenge is how to incorporate pseudo spurious labels into training in a principled manner so that spurious correlations are explicitly reduced at the feature level. 
In a typical unsupervised debiasing pipeline, one first trains a classifier to infer spurious attributes or environments from biased data and then uses the inferred spurious information to guide mitigation while training the target task classifier~\cite{liu2021justtraintwiceimproving,pezeshki2024discoveringenvironmentsxrm}.
As illustrated in Fig.~\ref{fig:intro}, most existing methods follow a single‑branch debiasing paradigm, where both the target task and spurious attribute prediction are modeled within a shared feature representation and a single network branch. Despite their effectiveness, single‑branch methods are inherently limited by the intrinsic entanglement between target and spurious attributes. Because both are encoded in the same feature space, features optimized for target prediction inevitably incorporate spurious cues, while features intended to model spurious attributes also retain class‑relevant information. This entanglement leads to suboptimal results on both target and spurious attribute prediction. 

To better isolate the target task and spurious attribute prediction, we introduce Dual‑branch Cross‑projection Debiasing (DCD), a two‑branch framework that explicitly separates their representation learning. Specifically, DCD maintains one prompt-tuned branch for the target task prediction and another for the spurious attribute prediction, and uses null-space projection to isolate the core information of each branch. Rather than balancing the spurious attribute only within the target feature space, DCD performs cross-projection between the two branches, filtering target features through the null space of the spurious branch and spurious features through the null space of the target branch. In this way, DCD suppresses spurious information while preserving target-relevant semantics, enabling bidirectional debiasing at the feature level. 

In conclusion, in this work, we propose Dual-Branch Cross-Projection Debiasing through Diffusion-based Disentanglement (D2CP), a unified framework for group-robust learning under spurious correlations. Our main contributions are threefold:
\begin{itemize} [leftmargin=*]
    \item We design Confidence-guided Bias Concept Mining (CBCM) to use diffusion-disentangled concepts for deriving semantic aligned pseudo spurious supervision in downstream debiasing. 
    
    \item We propose Dual-branch Cross-projection Debiasing (DCD), a dual-branch prompt-tuning based debiasing framework that explicitly separates their representation learning of spurious attributes and target attributes. 
    
    \item Extensive experiments on four benchmarks show that D2CP achieves strong worst-group performance among group unsupervised methods by tuning a very small fraction of parameters.
\end{itemize}
    
    


\section{Related Work}

\textbf{Improving group robustness. }
Existing methods for improving group robustness can be broadly categorized into two settings, depending on whether group annotations are available during training. (i) Group-supervised methods assume access to group labels and typically improve robustness by balancing the contribution of different groups, for example through sample reweighting, resampling~\cite{idrissi2022simpledatabalancingachieves,izmailov2022featurelearningpresencespurious,jung2022learningfairclassifierspartially,kirichenko2023layerretrainingsufficientrobustness}, robust optimization~\cite{arjovsky2020invariantriskminimization,sagawa2020distributionallyrobustneuralnetworks} or contrastive learning~\cite{park2022faircontrastivelearningfacial} . (ii) Group-unsupervised methods operate without group annotations. A common strategy in this setting is to first train a biased ERM model and then use its predictions or confidence patterns to infer pseudo-group labels, which are subsequently used to guide debiasing via sample rebalancing~\cite{liu2021justtraintwiceimproving,nam2020learningfailuretrainingdebiased,qiu2023simplefastgrouprobustness}, robust optimization~\cite{creager2021environmentinferenceinvariantlearning,nam2022spreadspuriousattributeimproving,sohoni2022subclassleftbehindfinegrained}. Another line of work improves robustness by synthesizing minority-group samples and using the generated data to train a more balanced and robust classifier~\cite{zhao2025aimfairadvancingalgorithmicfairness}.

\textbf{Parameter-Efficient Fine-Tuning. }
Conventional fine-tuning adapts downstream tasks by updating all model parameters,  but prior studies have shown that comparable performance can often be achieved by optimizing only a small set of additional parameters or a subset of the original parameters~\cite{han2024parameter,he2023sensitivity,he2021towards}. A representative early approach is adapter~\cite{adapter,he2021towards,chen2022adaptformer,sung2022vl,StPR,IKI,RD-MLDG,PAPT}, which inserts lightweight bottleneck modules between layers of a frozen backbone to capture task-specific information. By keeping the backbone fixed, adapters significantly reduce training cost while maintaining strong transferability~\cite{he2026harnessing}. Another widely adopted method is LoRA~\cite{aghajanyan2021intrinsic,edalati2025krona,dermino}, which parameterizes weight updates as low-rank decompositions~\cite{he2026taskdrivensubspacedecompositionknowledge}, enabling efficient adaptation with only a small number of trainable parameters. In addition, prompt tuning~\cite{PromptTuning, PrefixTuning,PADG,DPR,IVPT,MGSA} has emerged as an appealing alternative, where a small set of learnable prompts guides the model to leverage its pre-trained knowledge for downstream adaptation with minimal parameter updates.

\begin{figure*}[t]
    \centering
    \includegraphics[width=\textwidth]{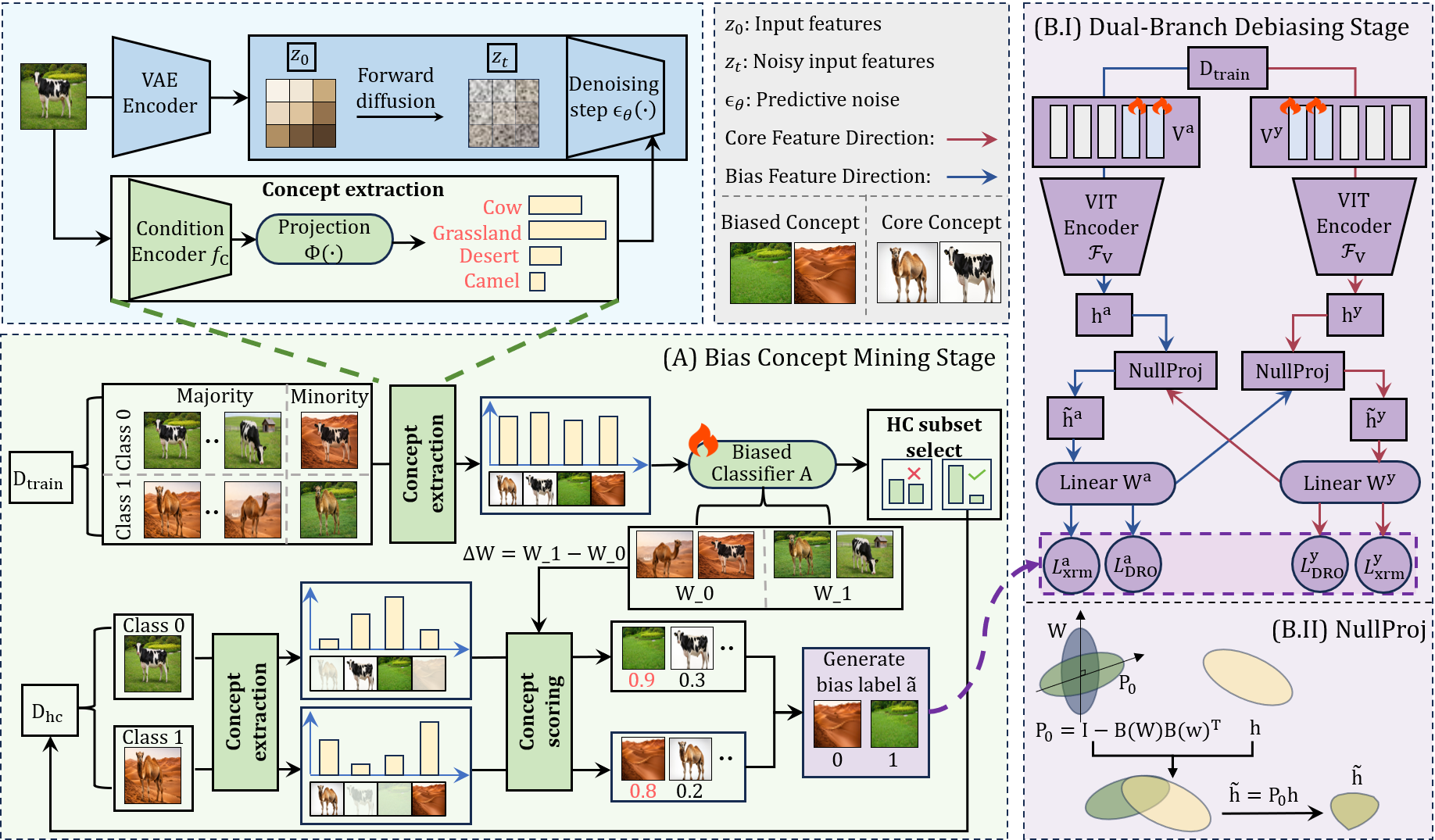}
\caption{Overview of the proposed framework. The framework consists of two stages: bias concept mining and dual-branch debiasing. In the first stage, a diffusion-based disentanglement model decomposes each image into concept tokens, and CBCM mines pseudo spurious labels from these concepts. In the second stage, DCD introduces two prompt-tuned branches and performs cross-projection to suppress spurious information while preserving class-relevant semantics. Finally, GroupDRO is used for downstream robust training.}
    \label{fig:framework}
\end{figure*}

\section{Methods}
\subsection{Problem Definition}

We study a multi-class prediction problem in which training data may contain spurious correlations between the target labels and spurious attributes. Each instance $x_i$ belongs to the input domain $\mathcal{X}$ (with $i=1,\ldots,N$ indexing the $N$ samples in the dataset) and is paired with a target label $y_i\in\mathcal{Y}=\{1,\ldots,Y\}$ as well as an attribute variable $a_i\in\mathcal{A}=\{1,\ldots,A\}$.  We partition the data into $|\mathcal{G}|=|\mathcal{Y}|\cdot|\mathcal{A}|$ subpopulations by the joint pair of label and attribute. Specifically, each group is indexed by $g=(y,a)\in\mathcal{G}\subseteq \mathcal{Y}\times\mathcal{A}$. Let $\mathcal{D}_{\text{train}}=\{(x_i,y_i)\}_{i=1}^{N}$ be the training set sampled i.i.d.\ from an unknown distribution $P$ over $\mathcal{X}\times\mathcal{Y}$, and denote by $P_g$ the conditional distribution restricted to group $g$.

In our setting, the spurious attribute $a$ is not observed during training. We aim to learn a neural network
$f_{\theta}:\mathcal{X}\rightarrow\mathbb{R}^{Y}$ that achieves strong overall performance while remaining robust across groups induced by the spurious attribute. Following prior work~\cite{sagawa2019distributionally}, we achieve group robustness by minimizing the worst-group risk:
\begin{equation}
\mathcal{R}_{\mathrm{wg}}(\theta)
=\max_{g\in\mathcal{G}}\ \mathbb{E}_{(x,y)\sim P_g}\!\left[\ell\big(y,f_{\theta}(x)\big)\right].
\end{equation}


\subsection{Overall framework}
Our framework consists of two stages, as illustrated in Fig.~\ref{fig:framework}. To mine pseudo spurious-attribute labels, we propose a Confidence-guided Bias Concept Mining (CBCM) module. 
Specifically, we first train a diffusion-based disentanglement model to decompose each image into multiple semantically meaningful concept tokens.
We then train a biased classifier on the raw training data.
Based on the classifier confidence, we select high-confidence samples within each class and aggregate their concept features to form class-specific prototypes.
Finally, we score these prototypes according to how strongly they activate the biased classifier, and select the highest-scoring ones as candidate spurious concepts.
These selected concepts are then converted into pseudo spurious labels for downstream debiasing.
To effectively leverage these pseudo labels, we further propose a Dual-branch Cross-projection Debiasing (DCD) module in the second stage.
DCD introduces two prompt-tuned branches over a shared frozen visual backbone, where one branch focuses on target prediction and the other captures spurious information. 
DCD designs a cross-projection training framework between the two branches to suppress spurious information while preserving class-relevant semantics.

\subsection{Preliminaries}

To obtain concepts suitable for spurious attribute mining, we build on EncDiff~\cite{Tao2024EncDiff}, a diffusion-based framework for learning disentangled visual concepts. Specifically, EncDiff encodes each image into a set of concept tokens, which are then used to condition image reconstruction in the latent diffusion model through cross-attention. This encourages the learned concept tokens to align with distinct spatial features during reconstruction, thereby facilitating the disentanglement of semantically meaningful concepts.

Given an image $\mathbf{x}_i$, EncDiff introduces a CNN-based condition encoder $f_C(\cdot)$ to extract a set of concept tokens from $\mathbf{x}_i$:
\begin{equation}
\mathbf{S}(\mathbf{x}_i)=f_C(\mathbf{x}_i)=\{\mathbf{s}_{i,1},\ldots,\mathbf{s}_{i,K}\}\in\mathbb{R}^{K},
\label{eq:weight_defination}
\end{equation}
where $\mathbf{S}(\mathbf{x}_i)$ denotes the concept set of image $\mathbf{x}_i$. $\mathbf{s}_{i,k}$ denotes the $k$-th concept token and $K$ is the number of concepts. A shared projection $\Phi(\cdot)$ then maps the concept tokens into the conditioning space of the denoising U-Net, so that they can be used as concept-level guidance for image reconstruction:
\begin{equation}
\boldsymbol{\Phi}(\mathbf{S}(\mathbf{x}_i))
=
\{\mathbf{E}_{i,1},\cdots,\mathbf{E}_{i,K}\}\in\mathbb{R}^{K\times D},
\label{eq:mapping_function}
\end{equation}
where $\mathbf{E}_{i,k}=\boldsymbol{\Phi}(\mathbf{s}_{i,k})\in\mathbb{R}^{D}$ denotes the projected feature of the $k$-th concept token and $D$ denotes the feature dimension.

Conditioned on the projected concept features, the denoising U-Net \(\boldsymbol{\epsilon}_{\theta}(\cdot)\) is trained to predict the injected noise into the latent representation at diffusion step $t$:
\begin{equation}
\mathcal{L}_{\mathrm{LDM}}
=
\mathbb{E}_{\mathbf{z}_0,t,\boldsymbol{\epsilon}\sim\mathcal{N}(\mathbf{0},\mathbf{I})}
\left[
\left\|
\boldsymbol{\epsilon}
-
\boldsymbol{\epsilon}_{\theta}\big(\mathbf{z}_t,t,\boldsymbol{\Phi}(\mathbf{S}(\mathbf{x}_i))\big)
\right\|_2^2
\right].
\end{equation}
Here, \(\mathbf{z}_0\) denotes the clean latent representation of image $\mathbf{x}_i$ obtained from the VAE encoder in the latent diffusion framework, and \(\mathbf{z}_t\) denotes its noisy version at diffusion step \(t\).
\(\boldsymbol{\epsilon}\) is the Gaussian noise added to \(\mathbf{z}_0\), while $\boldsymbol{\epsilon}_{\theta}\big(\mathbf{z}_t,t,\boldsymbol{\Phi}(\mathbf{S}(\mathbf{x}_i))\big)$ is the noise predicted by the U-Net from the noisy latent \(\mathbf{z}_t\) conditioned on $\boldsymbol{\Phi}(\mathbf{S}(\mathbf{x}_i))$.

By training the diffusion model and the condition encoder, we decompose each image into multiple disentangled concept-level factors. Building on these concepts, we next
introduce the CBCM module to mine candidate spurious concepts
for debiasing.

\subsection{Confidence-guided Bias Concept Mining (CBCM)}
To identify bias-related concepts from the diffusion-disentangled concept set, we propose a Confidence-guided Bias Concept Mining (CBCM) module. Our key intuition is that, when trained on biased data, a classifier tends to make predictions based on easy-to-learn bias-related cues, and these cues are more likely to be associated with spurious attributes. Based on this idea, CBCM consists of three steps: (1) training a biased classifier on the raw training data; (2) using this classifier to select high-confidence samples that are more likely to be dominated by bias-related cues, and aggregating their concepts within each class to form class prototypes, each representing a characteristic concept of these samples; and (3) scoring each prototype based on how strongly it activates the biased classifier to mine candidate spurious concepts for downstream debiasing.

\textbf{Biased classifier training.}
For each image-label pair $(\mathbf{x}_i,y_i)$ sampled from the biased training data $\mathcal{D}_{train}$, we concatenate its concept feature $\boldsymbol{\Phi} (\mathbf{S}(\mathbf{x}_i))$ to form an image-level representation:
\begin{equation}
\boldsymbol{\phi}(\mathbf{x}_i)=[\,\mathbf{E}_{i,1};\mathbf{E}_{i,2};\cdots;\mathbf{E}_{i,K}\,]\in\mathbb{R}^{KD},
\end{equation}
where $[\cdot;\cdot]$ denotes concatenation along the feature dimension. 
Based on the resulting image-level representation $\boldsymbol{\phi}(\mathbf{x}_i)$, we train a lightweight linear classifier $\mathbf{A}$ while keeping both $\mathbf{f}_C(\cdot)$ and $\boldsymbol{\Phi}(\cdot)$ frozen. The classifier output $\mathbf{z}^A$ is given by:
\begin{equation}
\mathbf{z}^A(\mathbf{x}_i)=\mathbf{W}^A\boldsymbol{\phi}(\mathbf{x}_i)+\mathbf{b}^A,
\end{equation}
where $\mathbf{W}^A\in\mathbb{R}^{2\times KD}$ and $\mathbf{b}^A\in\mathbb{R}^{2}$. For concept-level analysis, we reshape $\mathbf{W}^A$ into $\widetilde{\mathbf{W}}^A\in\mathbb{R}^{2\times K\times D}$, where $\widetilde{\mathbf{W}}^A_{c,k,:}\in\mathbb{R}^{D}$ denotes the classifier weight associated with concept $k$ for class $c\in\{0,1\}$.
The classifier $\mathbf{A}$ is trained using the standard cross-entropy loss:
\begin{equation}
\begin{aligned}
\mathcal{L}_{\mathrm{CE}}^{A}=
-&\frac{1}{|\mathcal{D}_{\mathrm{train}}|}
\sum_{(\mathbf{x}_i,y_i)\in\mathcal{D}_{\mathrm{train}}}
\log p_A(y_i \mid \mathbf{x}_i), \\
\text{where}\quad
&p_A(y_i \mid \mathbf{x}_i)
=
\frac{\exp(\mathbf{z}^A_{y_i}(\mathbf{x}_i))}
{\sum_{c\in\{0,1\}} \exp(\mathbf{z}^A_{c}(\mathbf{x}_i))},
\end{aligned}
\end{equation}
Here, $p_A(y_i \mid \mathbf{x}_i)$ denotes the probability assigned by classifier $\mathbf{A}$ to the ground-truth class $y_i$ for image $\mathbf{x}_i$, which is later used as the confidence score for within-class sample ranking.

\textbf{High-confidence subset selection.}
To reduce noise before prototype aggregation, we construct a class-balanced high-confidence subset by retaining confidently classified samples within each class. Specifically, for each class $c\in\{0,1\}$, let:
\begin{equation}
\mathcal{I}_c=\{\,i \mid (\mathbf{x}_i,y_i)\in\mathcal{D}_{\mathrm{train}},\; y_i=c\,\},
\end{equation}
denote the index set of samples belonging to class $c$. We rank the samples in $\mathcal{I}_c$ according to $p_A(y_i \mid \mathbf{x}_i)$ and retain the top $\rho\%$, yielding the selected index set $\mathcal{I}_c^{\mathrm{hc}}\subseteq\mathcal{I}_c$. The resulting class-balanced high-confidence subset $\mathcal{D}_{\mathrm{hc}}$ is defined as:
\begin{equation}
\mathcal{D}_{\mathrm{hc}}
=
\{\,(\mathbf{x}_i,y_i)\in\mathcal{D}_{\mathrm{train}} \mid i\in\mathcal{I}_0^{\mathrm{hc}}\cup \mathcal{I}_1^{\mathrm{hc}}\,\}.
\label{eq:cleaner_subset}
\end{equation}
Based on the retained high-confidence samples in each class, we construct a class-specific prototype for each concept. For each class $c\in\{0,1\}$ and concept index $k\in\{1,\cdots,K\}$, we compute the average projected concept feature over the retained high-confidence samples:
\begin{equation}
\bar{\mathbf{E}}_{c,k}
=
\frac{1}{\left|\mathcal{I}_c^{\mathrm{hc}}\right|}
\sum_{i \in \mathcal{I}_c^{\mathrm{hc}}} \mathbf{E}_{i,k}
\in \mathbb{R}^{D}.
\label{eq:bias_slot_proto}
\end{equation}
Here, $\bar{\mathbf{E}}_{c,k}$ denotes the class-specific prototype of concept $k$ under class $c$. 

\textbf{Spurious concept scoring.}
To characterize the concept-wise discriminative direction learned by classifier $\mathbf{A}$, we define:
\begin{equation}
\Delta \mathbf{W}
=
\widetilde{\mathbf{W}}^A_{1,:,:} - \widetilde{\mathbf{W}}^A_{0,:,:}
\in \mathbb{R}^{K \times D},
\label{eq:delta_w}
\end{equation}
where $\Delta \mathbf{W}_{k,:} \in \mathbb{R}^{D}$ denotes the discriminative direction associated with the $k$-th concept. We then assign a bias score to each concept by measuring the alignment between the class-specific concept prototype and the corresponding discriminative direction:
\begin{equation}
r_{c,k}
=
\left|
\left\langle
\bar{\mathbf{E}}_{c,k}, \Delta \mathbf{W}_{k,:}
\right\rangle
\right|.
\label{eq:bias_score}
\end{equation}
Here, $\langle \cdot,\cdot \rangle$ denotes the inner product between two $D$-dimensional feature. A larger score of $r_{c,k}$ indicates that the $k$-th concept is more strongly aligned with the classifier-derived label-discriminative direction, and is therefore more likely to capture a bias-related cue. For each class, we select the top-$M$ concept according to $\{r_{c,k}\}_{k=1}^{K}$:
\begin{equation}
\mathcal{T}_c = \mathrm{Top-M}(\{r_{c,k}\}_{k=1}^{K}),
\label{eq:topm_concepts}
\end{equation}
and define the final candidate spurious concept set $\mathcal{T}$ as:
\begin{equation}
\mathcal{T} = \mathcal{T}_0 \cap \mathcal{T}_1.
\label{eq:bias_concept_set}
\end{equation}

\textbf{Pseudo spurious label generation.}
Given the candidate spurious concept set $\mathcal{T}$, we derive a sample-level pseudo spurious label $\tilde{a}_i$ for each training sample to support downstream debiasing. Specifically, we compute a spurious score by aggregating the weights of the concepts indexed by $\mathcal{T}$, and assign $\tilde{a}_i$ by thresholding this score. In this way, the mined concept indices are converted into explicit pseudo supervision for subsequent debiasing.

\subsection{Dual-branch Cross-projection Debiasing (DCD)}

Based on the candidate spurious concepts identified by CBCM, we derive pseudo spurious labels to guide debiasing. Unlike existing single-branch debiasing methods, we propose a Dual-branch Cross-projection Debiasing (DCD) module. Our key intuition is that directly learning spurious features for debiasing is insufficient, since the learned spurious features may still encode class-relevant semantic information. As a result, debiasing based solely on such features may inadvertently remove knowledge essential for class prediction. To address this issue, DCD adopts a cross-projection training framework between the two branches, aiming to suppress spurious information while preserving class-relevant knowledge.


\textbf{Disentangled prompt tuning setting.} To disentangle class-relevant semantics from spurious attributes, we introduce two sets of learnable prompts $\mathbf{V}^{\mathrm{y}}, \mathbf{V}^{\mathrm{a}} \in \mathbb{R}^{L\times D_{\mathrm{mid}}}$, where $L$ denotes the prompt length and $D_{\mathrm{mid}}$ denotes the token dimension. Specifically, $\mathbf{V}^{\mathrm{y}}$ is designed to capture semantic information for class prediction, while $\mathbf{V}^{\mathrm{a}}$ is designed to capture spurious-related information for debiasing. Following VPT~\cite{jia2022visual}, we insert the learnable prompts into the token sequence of the frozen visual encoder $\mathcal{F}_{\mathrm{V}}(\cdot)$. Given an image $\mathbf{x}_i$, the feature representations of the two branches are obtained as:
\begin{equation}
    \mathbf{h}^{\mathrm{y}}_i=\mathcal{F}_{\mathrm{V}}(\mathbf{x}_i,\mathbf{V}^{\mathrm{y}}), \quad 
    \mathbf{h}^{\mathrm{a}}_i=\mathcal{F}_{\mathrm{V}}(\mathbf{x}_i,\mathbf{V}^{\mathrm{a}}).
\end{equation}
The two branch features, $\mathbf{h}^{\mathrm{y}}_i,\mathbf{h}^{\mathrm{a}}_i \in \mathbb{R}^{D_{\mathrm{mid}}}$, are not directly used for prediction; instead, they are first processed by cross null-space projection to obtain $\tilde{\mathbf{h}}^{\mathrm{y}}_i$ and $\tilde{\mathbf{h}}^{\mathrm{a}}_i$, which are then fed into the corresponding linear classifiers $\mathbf{W}^y,\mathbf{W}^a \in \mathbb{R}^{2\times D_{\mathrm{mid}}}$ for class and spurious prediction, respectively.


\textbf{Classifier-induced cross-projection.}
Before obtaining the final logits, we first construct branch-specific null-space projectors from the current classifiers to debias the learned features. For a binary classifier with weight matrix $\mathbf{W}\in\mathbb{R}^{2\times D_{\mathrm{mid}}}$, we compute an orthonormal basis $\mathbf{B}(\mathbf{W})\in\mathbb{R}^{D_{\mathrm{mid}}\times r}$ for the row space of $\mathbf{W}$, where $r=\mathrm{rank}(\mathbf{W})\leq 2$. The corresponding null-space projector is then defined as:
\begin{equation}
\mathbf{P}_{0}(\mathbf{W})=\mathbf{I}-\mathbf{B}(\mathbf{W})\mathbf{B}(\mathbf{W})^{\top}\in\mathbb{R}^{D_{\mathrm{mid}}\times D_{\mathrm{mid}}}.
\end{equation}
Using the semantic head $\mathbf{W}^{y}$ and the spurious head $\mathbf{W}^{a}$, we obtain two branch-specific projectors, denoted by $\mathbf{P}_0^{y}$ and $\mathbf{P}_0^{a}$, respectively.
These projectors are updated during each iteration and are subsequently used for cross-projection debiasing.


\textbf{Cross-projection debiasing.}
Given a mini-batch $\{(\mathbf{x}_i,y_i,\tilde{a}_i)\}_{i=1}^{B}$, we first extract target and spurious features $\mathbf{h}^{\mathrm{y}}_i$ and $\mathbf{h}^{\mathrm{a}}_i$ from the two branches. To further disentangle the learned feature representations, we then perform cross null-space projection using the previously defined projectors:
\begin{equation}
\tilde{\mathbf{h}}^{\mathrm{y}}_i=\mathbf{P}_0^{a}\mathbf{h}^{\mathrm{y}}_i,\qquad
\tilde{\mathbf{h}}^{\mathrm{a}}_i=\mathbf{P}_0^{y}\mathbf{h}^{\mathrm{a}}_i.
\end{equation}
The projected features are then fed into the corresponding classifiers for class and spurious prediction, respectively. For each image $\mathbf{x}_i$, we compute the cross-entropy losses $\ell_i^y$ and $\ell_i^a$ for the target and spurious branches against the class label $y_i$ and the pseudo spurious label $\tilde{a}_i$, respectively: 
\begin{equation}
\ell_i^{y}=\mathcal{L}_{\mathrm{CE}}(\mathbf{W}^{y}\tilde{\mathbf{h}}^{\mathrm{y}}_i,y_i),\qquad
\ell_i^{a}=\mathcal{L}_{\mathrm{CE}}(\mathbf{W}^{a}\tilde{\mathbf{h}}^{\mathrm{a}}_i,\tilde a_i).
\end{equation}


To further improve robustness, we extend the above sample-wise losses to the group level by applying GroupDRO\cite{sagawa2019distributionally}. Specifically, based on the class label $y_i$ and the pseudo spurious label $\tilde{a}_i$, we define the group index for image $\mathbf{x}_i$ as $g_i = 2y_i + \tilde{a}_i \in \{0,1,2,3\}$. For each branch, let $\ell_i\in\{\ell_i^y,\ell_i^a\}$ denote its corresponding sample-wise loss. Let $\mathcal{B}_g=\{\,i \mid g_i=g\,\}$ denote the set of images belonging to group $g$. The GroupDRO objective is then defined as:
\begin{equation}
\begin{aligned}
\mathcal{L}_g^y=
\frac{1}{|\mathcal{B}_g|}
\sum_{i\in\mathcal{B}_g}\ell_i^y,\quad
\mathcal{L}^y_{\mathrm{DRO}}
=
\sum_{g=0}^{3} q^y_g\mathcal{L}^y_g,
\\ 
\mathcal{L}^a_g=
\frac{1}{|\mathcal{B}_g|}
\sum_{i\in\mathcal{B}_g}\ell_i^a,\quad
\mathcal{L}^a_{\mathrm{DRO}}
=
\sum_{g=0}^{3} q^a_g\mathcal{L}^a_g,
\end{aligned}
\end{equation}
where $q_g \in \{q^y_g,q^a_g\}$ denotes the weight assigned to group-level loss $\mathcal{L}_g \in \{\mathcal{L}^y_g,\mathcal{L}^a_g\}$, with $\sum_{g=0}^{3}q_g=1$. Following GroupDRO~\cite{sagawa2019distributionally}, the group weights are updated during training as:
\begin{equation}
q_g
\leftarrow
\frac{q_g\exp(\eta L_g)}
{\sum_{g'=0}^{3} q_{g'}\exp(\eta L_{g'})},
\label{eq:weight_update}
\end{equation}
where $\eta>0$ is a hyperparameter controlling the strength of group reweighting. Larger group losses lead to larger weights, thereby improving worst-group robustness. 

To provide complementary supervision beyond the class-spurious grouping above, we further leverage XRM~\cite{pezeshki2024discoveringenvironmentsxrm} to generate auxiliary pseudo labels for training. Specifically, let $y_i^{\mathrm{xrm}}$ and $a_i^{\mathrm{xrm}}$ denote the XRM-generated pseudo labels for image $\mathbf{x}_i$, corresponding to the target and spurious branches, respectively. These pseudo labels are used to construct branch-specific auxiliary groupings that capture latent subgroup structure. Concretely, the auxiliary group index for the target branch is defined as $\gamma_i^{y}=2y_i+y_i^{\mathrm{xrm}}$, whereas that for the spurious branch is defined as $\gamma_i^{a}=2\tilde a_i+a_i^{\mathrm{xrm}}$. Let $\mathcal{B}_g^{y}=\{\,i \mid \gamma_i^{y}=g\,\}$ and $\mathcal{B}_g^{a}=\{\,i \mid \gamma_i^{a}=g\,\}$ denote the corresponding sample sets for the two branch-specific groupings. Accordingly, the auxiliary GroupDRO losses are defined as:
\begin{equation}
\begin{aligned}
\hat{\mathcal{L}}_g^{y}
&=
\frac{1}{|\mathcal{B}_g^{y}|}
\sum_{i\in\mathcal{B}_g^{y}} \ell_i^{y},
\quad
\mathcal{L}_{\mathrm{xrm}}^{y}
=
\sum_{g=0}^{3} \hat{q}_g^{y}\hat{\mathcal{L}}_g^{y},
\\
\hat{\mathcal{L}}_g^{a}
&=
\frac{1}{|\mathcal{B}_g^{a}|}
\sum_{i\in\mathcal{B}_g^{a}} \ell_i^{a},
\quad
\mathcal{L}_{\mathrm{xrm}}^{a}
=
\sum_{g=0}^{3} \hat{q}_g^{a}\hat{\mathcal{L}}_g^{a},
\end{aligned}
\end{equation}
where $\hat{q}_g^y,\hat{q}_g^a$ are updated in the same manner as Eq.~\ref{eq:weight_update}. The total training objective is defined as:
\begin{equation}
\mathcal{L}_{\mathrm{total}}=\underbrace{\mathcal{L}^y_{\mathrm{DRO}}+\mathcal{L}_{\mathrm{xrm}}^{y}}_{\textbf{Target Branch}}+\underbrace{\mathcal{L}^a_{\mathrm{DRO}}+\mathcal{L}_{\mathrm{xrm}}^{a}}_{\textbf{Spurious Branch}}.
\end{equation}


\section{Experiments}
In this section, we evaluate the effectiveness of our method on four benchmark datasets in the presence of spurious features, including Waterbirds\cite{Waterbirds}, CelebA\cite{celeba}, MetaShift\cite{metashift}, and CMNIST\cite{Cmnist}. Additional experimental details are provided in Appendix A.

\begin{table*}[t]
\centering
\small
\setlength{\tabcolsep}{6pt}
\renewcommand{\arraystretch}{1.0}
\caption{Results on four benchmarks. Best results are shown in bold. Test performance is reported using the checkpoint that achieves the highest validation worst-group accuracy. All results are averaged over three random seeds, with standard deviations reported. ‘Yes/No’ indicates whether group labels are used. ‘-’ denotes unavailable results. ‘*’ indicates results obtained from our training.}

\begin{tabular}{lllcccccccc}
\toprule
\textbf{Methods} & \textbf{Group} & \textbf{Venue} & \multicolumn{2}{c}{\textbf{Waterbirds}} & \multicolumn{2}{c}{\textbf{CelebA}} & \multicolumn{2}{c}{\textbf{MetaShift}} & \multicolumn{2}{c}{\textbf{CMNIST}} \\
\cmidrule(lr){4-5} \cmidrule(lr){6-7} \cmidrule(lr){8-9} \cmidrule(lr){10-11}
& \textbf{Info} &  & WGA(\%)$\uparrow$ & Mean(\%) & WGA(\%)$\uparrow$ & Mean(\%) & WGA(\%)$\uparrow$ & Mean(\%) & WGA(\%)$\uparrow$ & Mean(\%) \\
\midrule
Group DRO\cite{sagawa2019distributionally} & Yes & ICLR 20            & 86.5 & 90.2 & 88.3 & 93.1 & 80.3 & -    & 88.3*{\scriptsize$\pm$4.5}     & 97.4*{\scriptsize$\pm$1.3}     \\
PDE\cite{PDE} & Yes & NeurIPS 23     & 90.3{\scriptsize$\pm$0.3} & 92.4{\scriptsize$\pm$0.8} & 91.0{\scriptsize$\pm$0.4} & 92.0{\scriptsize$\pm$0.6} & 77.43*{\scriptsize$\pm$0.89}    & 87.6*{\scriptsize$\pm$3.07}    & 90.73*{\scriptsize$\pm$2.39}     & 98.16*{\scriptsize$\pm$0.24}    \\
CPT\cite{CPT} & Yes & ICML 24      & 93.5{\scriptsize$\pm$0.4} & 96.3{\scriptsize$\pm$0.2} & 92.0{\scriptsize$\pm$0.3} & 93.2{\scriptsize$\pm$0.3} & 83.2*{\scriptsize$\pm$1.46}    & 90.24*{\scriptsize$\pm$0.17}    & 91.3*{\scriptsize$\pm$3.77}     & 97.85*{\scriptsize$\pm$1.25}    \\

PDE+EIRep\cite{EIRep} & Yes & AISTATS 25 & 90.4{\scriptsize$\pm$0.2} & 91.6{\scriptsize$\pm$0.7} & 91.4{\scriptsize$\pm$0.5} & 92.4{\scriptsize$\pm$0.3} & -    & -    & -     & -    \\
CRM\cite{CRM} & Yes & ICML 25         & 86.0{\scriptsize$\pm$0.6} & -    & 89.0{\scriptsize$\pm$0.6} & -    & 74.7{\scriptsize$\pm$1.5} & -    & 94.3*{\scriptsize$\pm$0.9}     & 99.0*{\scriptsize$\pm$0.0}     \\
\midrule
JTT\cite{liu2021justtraintwiceimproving} & No & ICML 21         & 86.7 & 93.3 & 81.1 & 88.0 & 69.6*    & 78.1*    & 69.6*     & 99.6*    \\
CNC\cite{cnc} & No & ICML 22         & 88.5{\scriptsize$\pm$0.3} & 90.9{\scriptsize$\pm$0.1} & 88.8{\scriptsize$\pm$0.9} & 89.9{\scriptsize$\pm$0.5} & -    & -    & -     & -    \\
C-Adapter\cite{C-Adapter} & No & NeurIPS 22         & 86.9 & 96.2 & 90.0 & 90.7 & -    & -    & -     & -    \\
MaskTune\cite{asgari2022masktune} & No & NeurIPS 22         & 86.4{\scriptsize$\pm$1.9} & 93.0{\scriptsize$\pm$0.7} & 78.0{\scriptsize$\pm$1.2} & 91.3{\scriptsize$\pm$0.1} & 72.31*{\scriptsize$\pm$6.71}    & 87.68*{\scriptsize$\pm$1.56}    & 2.17*{\scriptsize$\pm$3.77}     & 52.49*{\scriptsize$\pm$2.02}    \\
uLA\cite{ula} & No & NeurIPS 23      & 86.1{\scriptsize$\pm$1.5} & 91.5{\scriptsize$\pm$0.7} & 86.5{\scriptsize$\pm$3.7} & 93.9{\scriptsize$\pm$0.2} & -    & -    & -     & -    \\
GIC\cite{GIC} & No & ICML 24      & 86.3{\scriptsize$\pm$0.1} & 89.6{\scriptsize$\pm$1.3} & 89.5{\scriptsize$\pm$0.0} & 92.1{\scriptsize$\pm$0.1} & -    & -    & -     & -    \\
XRM\cite{pezeshki2024discoveringenvironmentsxrm} & No & ICML 24         & 88.1{\scriptsize$\pm$0.9} & 89.3 & 89.1{\scriptsize$\pm$1.3} & 91.4 & 77.5{\scriptsize$\pm$1.0} & -    & 90.5*{\scriptsize$\pm$2.1} & 98.0*{\scriptsize$\pm$0.5}    \\

SAS\cite{SAS} & No & ICLR 25         & 89.7 & 96.3 & 87.4 & 91.1 & -    & -    & -     & -    \\

PPA\cite{zhu2025projectprobeaggregateefficientfinetuninggroup} & No & CVPR 25         & 87.2 & 94.6 & 91.1 & 92.1 & -    & -    & -     & -    \\

GERNE\cite{Gerne} & No & ICCV 25       & 89.88{\scriptsize$\pm$0.67} & -   & 74.24{\scriptsize$\pm$2.51} & -   & -    & -    & -     & -    \\
D2CP(ours) & No &        & \textbf{90.81}{\scriptsize$\pm$0.31} & 93.06{\scriptsize$\pm$0.41}   & \textbf{91.80}{\scriptsize$\pm$0.27} &  92.34{\scriptsize$\pm$0.33}  & \textbf{82.20}{\scriptsize$\pm$1.13}    &  88.25{\scriptsize$\pm$1.08}   & \textbf{91.99}{\scriptsize$\pm$1.20}     &  96.39{\scriptsize$\pm$1.19}   \\
\bottomrule
\end{tabular}

\label{tab:main_results}
\end{table*}

\subsection{Datasets}
\textbf{Waterbirds}~\cite{Waterbirds} is constructed by compositing bird foregrounds from CUB\cite{CUB} with background scenes from Places\cite{zhou2017places}. The dataset contains 11,788 images and exhibits a strong correlation between bird category and scene context. We treat the background type, namely water or land, as the spurious attribute, and the bird category, namely waterbird or landbird, as the core attribute.

\textbf{CelebA}~\cite{celeba} consists of 202,599 face images annotated with 40 binary attributes. The dataset exhibits a strong correlation between hair color and gender: most blond-hair samples are female, whereas most non-blond-hair samples are male. We treat gender, namely male or female, as the spurious attribute, and hair color, namely blond hair or non-blond hair, as the core attribute.

\textbf{MetaShift}~\cite{metashift} is a binary classification dataset of 3,499 cat and dog images. Following \cite{pezeshki2024discoveringenvironmentsxrm}, we consider the standard spurious-correlation setting where cats are predominantly associated with indoor backgrounds and dogs with outdoor backgrounds. In this setting, background type (indoor vs.\ outdoor) is treated as the spurious attribute, while object category (cat vs.\ dog) is treated as the core attribute.


\textbf{C-MNIST}~\cite{Cmnist} is a synthetic MNIST variant with 70,000 digit images. Following \cite{chakraborty2024exmap}, we formulate it as a binary classification task by grouping digits 0--4 and 5--9 into two classes. To introduce spurious correlations, each grayscale digit is converted into a three-channel colored image, where red and green denote the spurious attribute. We regard color as the spurious attribute and digit class as the core attribute.

\subsection{Baselines}

Following prior work, we compare our method against a diverse set of representative baselines for mitigating spurious correlations. Specifically, we include methods that use group supervision during training or model selection, including CPT\cite{CPT}, CRM\cite{CRM}, Group DRO\cite{sagawa2020distributionallyrobustneuralnetworks}, PDE\cite{PDE}, and PDE+EIRep\cite{EIRep}. We also compare with a broad set of methods that do not rely on group annotations, including JTT\cite{liu2021justtraintwiceimproving}, CNC\cite{cnc}, C-Adapter\cite{C-Adapter}, MaskTune\cite{asgari2022masktune}, uLA\cite{ula}, GIC\cite{GIC}, XRM\cite{pezeshki2024discoveringenvironmentsxrm} , SAS\cite{SAS}, PPA\cite{zhu2025projectprobeaggregateefficientfinetuninggroup}, and GERNE\cite{Gerne}.

\subsection{Training and Evaluation Details}

Following~\cite{CPT}, We use ViT-B/16 as the backbone and adapt prompt tuning for all experiments. The prompt length is set to $L{=}20$ for CelebA~\cite{celeba} and $L{=}10$ for CMNIST~\cite{Cmnist}, MetaShift~\cite{metashift}, and Waterbirds~\cite{Waterbirds}. We train for 10 epochs on CMNIST and 20 epochs on all other datasets. SGD is used as the optimizer throughout. We apply StepLR on CelebA and MetaShift, while no learning rate scheduler is used for Waterbirds or CMNIST. Additional implementation details are provided in Appendix~B.

For evaluation, we report worst-group accuracy and mean accuracy on all datasets. To ensure a fair comparison with prior work, we directly adapt baseline results from the original papers whenever available, rather than re-implementing and re-running all competing methods.

\subsection{Main Results}

Table~\ref{tab:main_results} summarizes the results on Waterbirds~\cite{Waterbirds}, CelebA~\cite{celeba}, MetaShift~\cite{metashift}, and CMNIST~\cite{Cmnist}. D2CP achieves the best worst-group accuracy (WGA) among all group-unsupervised methods on all four benchmarks, while remaining competitive in mean accuracy.

\begin{table*}[t]
\centering
\small
\setlength{\tabcolsep}{8pt}
\renewcommand{\arraystretch}{1.15}
\caption{Main component analysis of D2CP. CBCM indicates whether pseudo labels derived from CBCM-based disentanglement are used; DCD indicates whether the dual-branch design with cross-projection is adopted; XRM\_aux indicates whether auxiliary group labels produced by XRM\cite{pezeshki2024discoveringenvironmentsxrm} are used.}
\begin{tabular}{c c c c | c c c c}
\hline
\textbf{Exp.} & \textbf{CBCM}  & \textbf{DCD} & \textbf{XRM\_aux} & \textbf{Waterbirds} & \textbf{CelebA} & \textbf{MetaShift} & \textbf{CMNIST} \\
\hline
(a) &  &  &  & 74.97{\scriptsize$\pm$2.78}  & 67.41{\scriptsize$\pm$4.49} & 71.65{\scriptsize$\pm$10.43} & 9.42{\scriptsize$\pm$14.48} \\
(b) & \ding{51} &  &  & 85.57{\scriptsize$\pm$2.53} & 84.44{\scriptsize$\pm$3.38} & 75.86{\scriptsize$\pm$5.18} & 87.20{\scriptsize$\pm$3.35} \\
(c) &  &  & \ding{51} & 86.29{\scriptsize$\pm$1.11} & 84.63{\scriptsize$\pm$1.61} & 74.85{\scriptsize$\pm$5.73} & 90.95{\scriptsize$\pm$3.82} \\
(d) & \ding{51} &  & \ding{51} & 86.88{\scriptsize$\pm$0.97} & 88.86{\scriptsize$\pm$1.95} & 81.63{\scriptsize$\pm$0.07} & 91.30{\scriptsize$\pm$2.18} \\
(e) & \ding{51} & \ding{51} &  & 83.28{\scriptsize$\pm$1.71} & 88.71{\scriptsize$\pm$2.74} & 71.80{\scriptsize$\pm$5.36} & 88.41{\scriptsize$\pm$1.25} \\
\hline
\rowcolor{gray!20}
(f) & \ding{51} & \ding{51} & \ding{51} & 90.81{\scriptsize$\pm$0.31} & 91.80{\scriptsize$\pm$0.27} & 82.20{\scriptsize$\pm$1.13} & 91.99{\scriptsize$\pm$1.20} \\
\hline
\end{tabular}

\label{tab:ablation}
\end{table*}

\noindent\textbf{Comparison with Group-Unsupervised Methods.}
D2CP attains 90.81\% WGA on Waterbirds, 91.80\% on CelebA, 82.20\% on MetaShift, and 91.99\% on CMNIST. Compared with the strongest prior group-unsupervised baselines, it improves WGA by 0.93, 0.70, 4.70, and 1.49 points on the four datasets, respectively. These consistent gains verify the effectiveness of D2CP for improving worst-group robustness without using group annotations.

\noindent\textbf{Comparison with Group-Supervised Methods.}
Although D2CP is fully group-unsupervised, it remains competitive with supervised methods that rely on group annotations. In particular, D2CP outperforms several supervised baselines on Waterbirds and CelebA, surpasses Group DRO, PDE and CRM on MetaShift, and exceeds Group DRO, PDE and CPT on CMNIST. These results suggest that D2CP can narrow the gap between group-unsupervised and group-supervised robust learning, and even outperform supervised approaches in several cases.

\noindent\textbf{Parameter Efficiency.}
D2CP is highly parameter-efficient. Using only 10 prompts per layer (20 for CelebA), it introduces approximately 94K trainable parameters, corresponding to just 0.11\% of ViT-B/16. Despite updating only these lightweight prompts together with a small classifier, while keeping the ViT-B/16 backbone frozen, D2CP still delivers strong worst-group robustness.

\subsection{Further Analysis and Ablation Studies}
We conduct comprehensive analyses to further examine D2CP from multiple perspectives. First, we evaluate the contribution of its key components, including Confidence-guided Bias Concept Mining (CBCM), Dual-branch Cross-projection Debiasing (DCD), and XRM-based auxiliary grouping (XRM\_aux), to understand how each module improves group robustness. 
Second, we quantitatively assess the quality of the pseudo spurious labels discovered by CBCM, both by comparing them with ground-truth spurious attributes and by analyzing their effectiveness as supervision for downstream debiasing. 
Third, we provide a qualitative analysis of the mined spurious concepts through concept manipulation and quantized reconstruction in diffusion generation, showing that the discovered concepts are semantically meaningful. 
Finally, we study the effect of the sampling exponent $\alpha$ in the weighted sampling strategy to better understand its impact on robust performance across datasets.


\textbf{The effectiveness of CBCM.}
As shown in Table~\ref{tab:ablation}, CBCM provides effective pseudo spurious supervision for debiasing. Compared with the ERM baseline in Row~(a), enabling CBCM alone in Row~(b) improves worst-group accuracy by 10.60, 17.03, 4.21, and 77.78 points on Waterbirds, CelebA, MetaShift, and CMNIST, respectively. CBCM remains beneficial when combined with XRM\_aux: comparing Row~(c) with Row~(d), it further improves worst-group accuracy by 0.59, 4.23, 6.78, and 0.35 points on the four datasets, respectively. These results indicate that the pseudo labels mined from diffusion-disentangled concepts provide reliable bias-related cues and consistently improve worst-group robustness.

\textbf{The effectiveness of DCD.}
We examine the role of DCD in Table~\ref{tab:ablation}. Comparing Row~(b) with Row~(e), adding DCD on top of CBCM improves worst-group accuracy by 4.27 points on CelebA and 1.21 points on CMNIST. More importantly, DCD yields consistent gains when combined with both CBCM and XRM\_aux. Specifically, comparing Row~(d) with Row~(f), introducing DCD further improves worst-group accuracy by 3.93, 2.94, 0.57, and 0.69 points on Waterbirds, CelebA, MetaShift, and CMNIST, respectively. These results suggest that DCD facilitates more effective debiasing through cross-projection between the target and spurious branches, which suppresses spurious information while retaining class-relevant semantics.

\begin{figure}[!t]
    \centering
    \includegraphics[width=0.48\textwidth,]{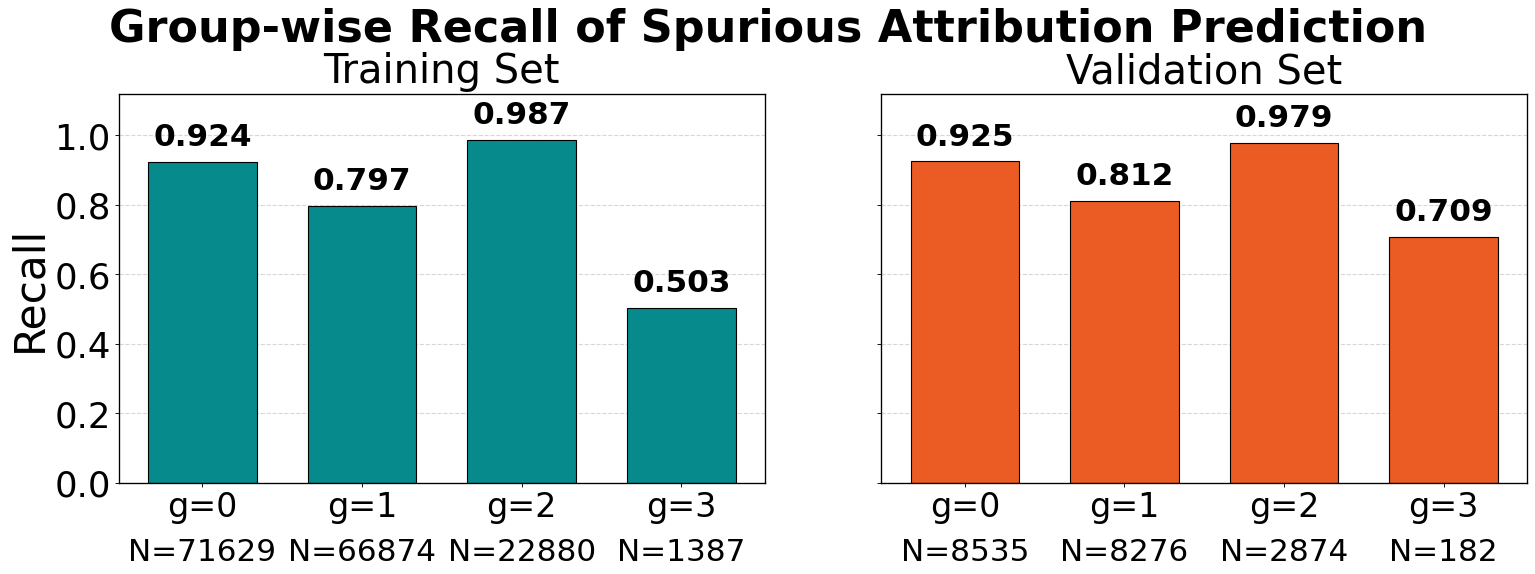}
    \caption{Group-wise recall of the pseudo spurious labels mined from diffusion-disentangled concepts on CelebA~\cite{celeba}, evaluated against the ground-truth spurious attribute labels. }
    \label{fig:prediction}
\end{figure}

\textbf{Quantitative evaluation of CBCM pseudo labels.}
To further assess the quality of the pseudo labels produced by CBCM, we compare the mined attribute assignments with the ground-truth spurious attribute labels across the four groups, indexed by \(g = 2y + a \in \{0,1,2,3\}\). As shown in Fig.~\ref{fig:prediction}, CBCM achieves high group-wise recall on most groups. On the training set, the recalls for \(g=0,1,2,3\) are \(92.4\%\), \(79.7\%\), \(98.7\%\), and \(50.3\%\), respectively, while the corresponding recalls on the validation set are \(92.5\%\), \(81.2\%\), \(97.9\%\), and \(70.9\%\). These results indicate that the pseudo labels mined by CBCM are largely consistent with the ground-truth spurious attribute labels, providing quantitative evidence that CBCM can discover meaningful spurious concepts and transform them into effective pseudo supervision for downstream debiasing.

\begin{figure*}[t]
    \centering
    \includegraphics[width=0.98\textwidth]{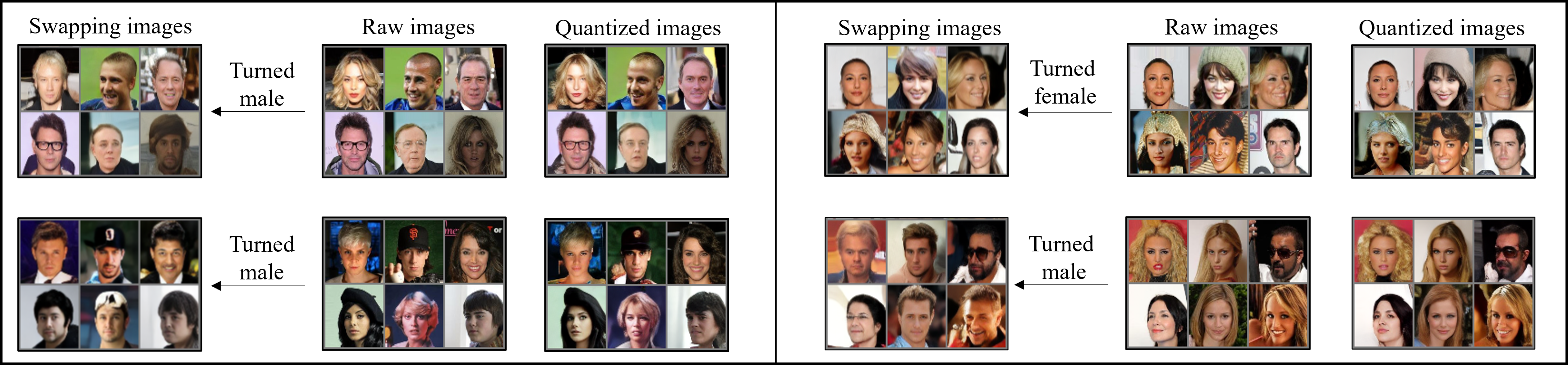}
    \caption{Qualitative analysis of concept manipulation and quantized reconstruction in diffusion generation. Middle: original images. Left: resampled images after replacing one condition dimension. Right: images decoded from the quantized predicted representation.}

    \label{fig:diffusion}
\end{figure*}

\textbf{Quality of CBCM-generated pseudo labels.}
Table~\ref{tab:label_compare} evaluates the quality of the pseudo spurious labels discovered by CBCM. 
For fair comparison, we keep exactly the same hyper-parameter settings as those used in the main results in Table~\ref{tab:main_results}.
Across Waterbirds, CelebA, and CMNIST, training DCD with random labels leads to worse worst-group accuracy, indicating that arbitrary supervision fails to provide meaningful bias cues for debiasing. 
In contrast, replacing random labels with CBCM-generated pseudo labels consistently brings large improvements on both worst-group and mean accuracy, showing that the mined concepts are strongly aligned with the underlying spurious attributes. 
Moreover, the performance obtained with CBCM labels is much closer to that achieved using ground-truth spurious labels across all three datasets, demonstrating that CBCM can recover reliable pseudo spurious labels that serve as effective supervision for downstream debiasing.

\textbf{Qualitative analysis of concept manipulation and quantized reconstruction.}
Fig.~\ref{fig:diffusion} shows results on CelebA~\cite{celeba} by manipulating the spurious concept discovered by CBCM and decoding the corresponding quantized predicted representation. Comparing the middle and left columns, replacing the selected condition dimension with that from another sample consistently changes the generated faces toward a different gender, while preserving most other visual details such as identity-related appearance and pose. This behavior suggests that the manipulated dimension corresponds to a semantically meaningful and controllable attribute, rather than causing arbitrary changes in the image. Comparing the middle and right columns, the quantized predicted representation can still be decoded into coherent images that preserve the main visual structure of the input. This observation indicates that the learned concept representation remains structured and semantically informative even after quantization. Taken together, these results provide qualitative evidence that CBCM can identify a meaningful and controllable spurious attribute.

\begin{table}[t!]
\centering
\small
\setlength{\tabcolsep}{6pt}
\renewcommand{\arraystretch}{1.2}
\caption{Comparison of different spurious-label settings for training DCD framework under the same hyper-parameter configuration as Table~\ref{tab:main_results}, illustrating the impact of different spurious supervision signals on debiasing performance.}

\resizebox{0.48\textwidth}{!}{%
\begin{tabular}{c c ccc}
\toprule
\textbf{Dataset} & \textbf{Metric} & \textbf{Random} & \textbf{CBCM} & \textbf{GT} \\
\midrule
\multirow{2}{*}{Waterbirds}
& WGA(\%)$\uparrow$ 
& 75.34{\scriptsize$\pm$4.26} 
& \cellcolor{gray!20}83.28{\scriptsize$\pm$1.71}
& 90.44{\scriptsize$\pm$0.49} \\
& Mean(\%) 
& 87.77{\scriptsize$\pm$2.33}
& \cellcolor{gray!20}90.28{\scriptsize$\pm$1.25}
& 92.65{\scriptsize$\pm$1.11} \\
\midrule

\multirow{2}{*}{CelebA}
& WGA(\%)$\uparrow$ 
& 68.51{\scriptsize$\pm$1.05}
& \cellcolor{gray!20}88.71{\scriptsize$\pm$2.74}
& 89.81{\scriptsize$\pm$1.40} \\
& Mean(\%) 
& 94.16{\scriptsize$\pm$0.15}
& \cellcolor{gray!20}92.97{\scriptsize$\pm$0.10}
& 93.42{\scriptsize$\pm$0.14} \\
\midrule

\multirow{2}{*}{CMNIST}
& WGA(\%)$\uparrow$ 
& 0.0{\scriptsize$\pm$0.0}
& \cellcolor{gray!20}88.41{\scriptsize$\pm$1.25}
& 92.75{\scriptsize$\pm$2.05} \\
& Mean(\%) 
& 99.06{\scriptsize$\pm$0.0}
& \cellcolor{gray!20}95.87{\scriptsize$\pm$0.83}
& 95.75{\scriptsize$\pm$1.04} \\
\bottomrule
\end{tabular}%
}
\label{tab:label_compare}
\end{table}

\begin{figure}[t]
    \centering
    \includegraphics[width=0.48\textwidth]{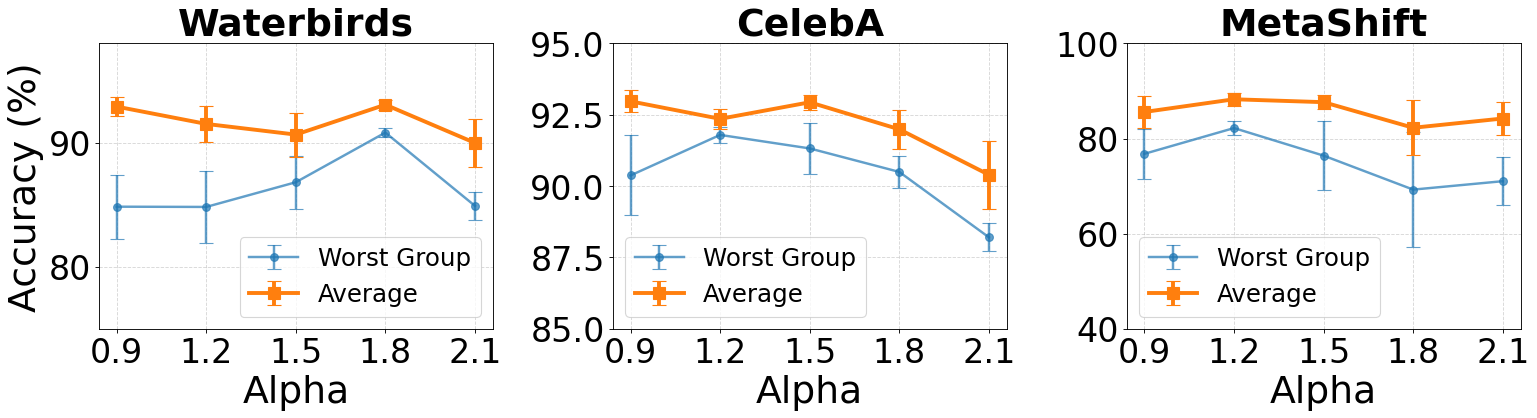}
    \caption{Effect of the sampling exponent $\alpha$ in the weighted sampling strategy across Waterbirds, CelebA, and MetaShift. Results are averaged over three random seeds, and the error bars denote the standard deviation across different runs.}
    \label{fig:hyper-parameter}
\end{figure}

\textbf{Qualitative analysis of the sampling exponent $\alpha$.}
We study the effect of the sampling exponent $\alpha$ in the weighted sampling strategy, as shown in Fig.~\ref{fig:hyper-parameter}. Specifically, we vary $\alpha$ from 0.9 to 2.1 and report both worst-group accuracy and average accuracy on Waterbirds~\cite{Waterbirds}, CelebA~\cite{celeba}, and MetaShift~\cite{metashift}. Larger $\alpha$ more strongly amplifies the sampling weights of minority groups, while $\alpha=1$ corresponds to the original weighting scheme. We find that increasing $\alpha$ generally improves worst-group performance in the low-$\alpha$ regime, but overly large values tend to harm training stability due to excessive over-sampling. Once the minority groups are over-emphasized, the model may overfit these samples and lose its ability to generalize well across all groups. Across all three datasets, the best worst-group accuracy is achieved at intermediate values of $\alpha$, with the optimal choices being around 1.8 for Waterbirds and 1.2 for both CelebA and MetaShift. These results indicate that moderate reweighting yields the best trade-off between minority-group robustness and overall accuracy. This observation highlights the importance of tuning $\alpha$ to match the bias severity and data distribution of each dataset.

\section{Conclusion}
In this paper, we proposed Dual-Branch Cross-Projection Debiasing through
Diffusion-based Disentanglement (D2CP), a unified framework for improving group robustness under spurious correlations without requiring group annotations during training. Our method leverages diffusion-based disentangled concept representations to discover label-correlated spurious concepts through Confidence-guided Bias Concept Mining (CBCM), which are converted into explicit pseudo spurious labels to provide bias priors for downstream debiasing. Building on these mined concepts, we further introduced Dual-branch Cross-projection Debiasing (DCD), a parameter-efficient prompt-tuning framework that suppresses spurious directions through cross null-space projection while preserving task-relevant semantics. Extensive experiments on multiple benchmark datasets show that D2CP consistently improves worst-group performance over prior group-unsupervised methods while maintaining strong average accuracy with only a small number of trainable parameters. This design enables D2CP to bridge concept-level bias discovery and downstream robust optimization in a unified and parameter-efficient manner.

\clearpage

\bibliographystyle{ACM-Reference-Format}
\bibliography{reference}


\appendix
\clearpage
\onecolumn

{\begin{center}
 \Large{Supplementary Materials for ``\textbf{Dual-Branch Cross-Projection Debiasing through Diffusion-based Disentanglement}''}   
\end{center}}

\section{Dataset statistics}

We conduct experiments with D2CP on Waterbirds~\cite{Waterbirds} (Table~\ref{tab:waterbirds_example}), CelebA~\cite{celeba} (Table~\ref{tab:celeba_example}), CMNIST~\cite{Cmnist} (Table~\ref{tab:cmnist_example}), and MetaShift~\cite{metashift} (Table~\ref{tab:metashift_example}), following established evaluation protocols~\cite{pezeshki2024discoveringenvironmentsxrm,chakraborty2024exmap}. These benchmarks involve standard image classification tasks; however, the target labels are typically entangled with spurious correlations. Detailed definitions of spurious attributes and group splits are provided in the corresponding tables.

\vspace{0.3em}

\begin{table}[!ht]
\centering
\scriptsize
\setlength{\tabcolsep}{4pt}
\renewcommand{\arraystretch}{1.05}
\caption{Example images of Waterbirds~\cite{sagawa2019distributionally}.}
\resizebox{0.82\textwidth}{!}{%
\begin{tabular}{c|c c c c}
\toprule
Image 
& \includegraphics[width=0.12\textwidth,height=1.0cm]{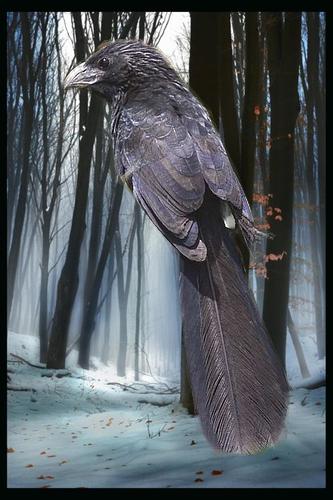}
& \includegraphics[width=0.12\textwidth,height=1.0cm]{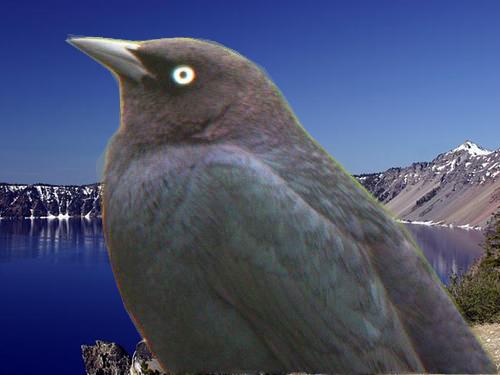}
& \includegraphics[width=0.12\textwidth,height=1.0cm]{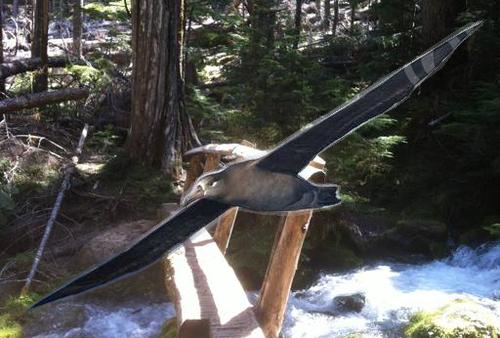}
& \includegraphics[width=0.12\textwidth,height=1.0cm]{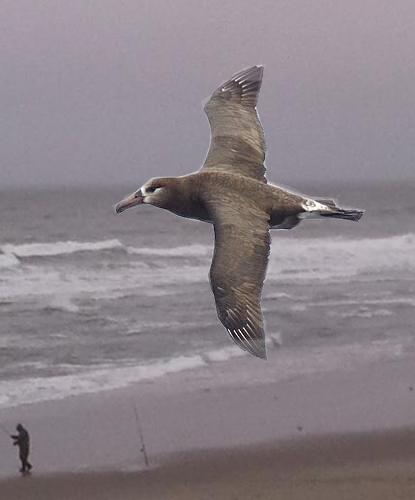} \\
\midrule
Label & 0 (landbird) & 0 (landbird) & 1 (waterbird) & 1 (waterbird) \\
Spurious feature & 0 (land) & 1 (water) & 0 (land) & 1 (water) \\
Train data & 3,498 (73\%) & 184 (4\%) & 56 (1\%) & 1,057 (22\%) \\
Val data & 467 & 466 & 133 & 133 \\
Test data & 2,255 & 2,255 & 642 & 642 \\
\bottomrule
\end{tabular}}
\label{tab:waterbirds_example}
\end{table}

\vspace{-1.6em}

\begin{table}[!ht]
\centering
\scriptsize
\setlength{\tabcolsep}{4pt}
\renewcommand{\arraystretch}{1.05}
\caption{Example images of CelebA~\cite{celeba}.}
\resizebox{0.82\textwidth}{!}{%
\begin{tabular}{c|c c c c}
\toprule
Image 
& \includegraphics[width=0.12\textwidth,height=1.0cm]{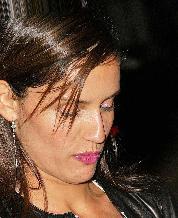}
& \includegraphics[width=0.12\textwidth,height=1.0cm]{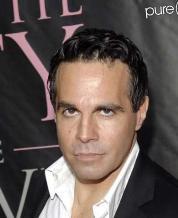}
& \includegraphics[width=0.12\textwidth,height=1.0cm]{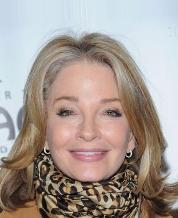}
& \includegraphics[width=0.12\textwidth,height=1.0cm]{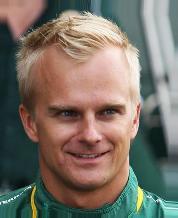} \\
\midrule
Label & 0 (non-blond) & 0 (non-blond) & 1 (blond) & 1 (blond) \\
Spurious feature & 0 (woman) & 1 (man) & 0 (woman) & 1 (man) \\
Train data & 71,629 (44\%) & 66,874 (41\%) & 22,880 (14\%) & 1,387 (1\%) \\
Val data & 8,535 & 8,276 & 2,874 & 182 \\
Test data & 9,767 & 7,535 & 2,480 & 180 \\
\bottomrule
\end{tabular}}
\label{tab:celeba_example}
\end{table}

\vspace{-1.6em}

\begin{table}[!ht]
\centering
\scriptsize
\setlength{\tabcolsep}{4pt}
\renewcommand{\arraystretch}{1.05}
\caption{Example images of CMNIST~\cite{Cmnist}.}
\resizebox{0.82\textwidth}{!}{%
\begin{tabular}{c|c c c c}
\toprule
Image 
& \includegraphics[width=0.12\textwidth,height=1.0cm]{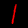}
& \includegraphics[width=0.12\textwidth,height=1.0cm]{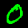}
& \includegraphics[width=0.12\textwidth,height=1.0cm]{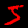}
& \includegraphics[width=0.12\textwidth,height=1.0cm]{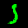} \\
\midrule
Label & 0 ($\texttt{value}\leq 4$) & 0 ($\texttt{value}\leq 4$) & 1 ($\texttt{value}>4$) & 1 ($\texttt{value}>4$) \\
Spurious feature & 0 (red) & 1 (green) & 0 (red) & 1 (green) \\
Train data & 254 (0.5\%) & 25,284 (50.6\%) & 24,231 (48.5\%) & 231 (0.5\%) \\
Val data & 45 & 5,013 & 4,893 & 49 \\
Test data & 48 & 5,091 & 4,815 & 46 \\
\bottomrule
\end{tabular}}
\label{tab:cmnist_example}
\end{table}

\vspace{-1.6em}

\begin{table}[!ht]
\centering
\scriptsize
\setlength{\tabcolsep}{4pt}
\renewcommand{\arraystretch}{1.05}
\caption{Example images of MetaShift~\cite{metashift}.}
\resizebox{0.82\textwidth}{!}{%
\begin{tabular}{c|c c c c}
\toprule
Image 
& \includegraphics[width=0.12\textwidth,height=1.0cm]{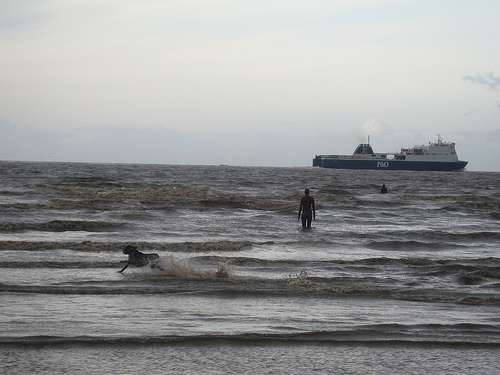}
& \includegraphics[width=0.12\textwidth,height=1.0cm]{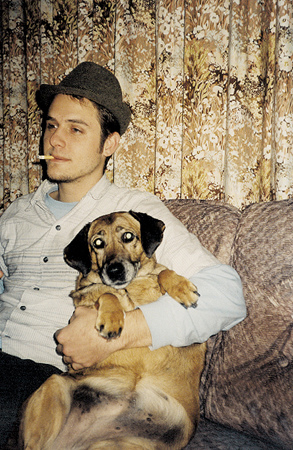}
& \includegraphics[width=0.12\textwidth,height=1.0cm]{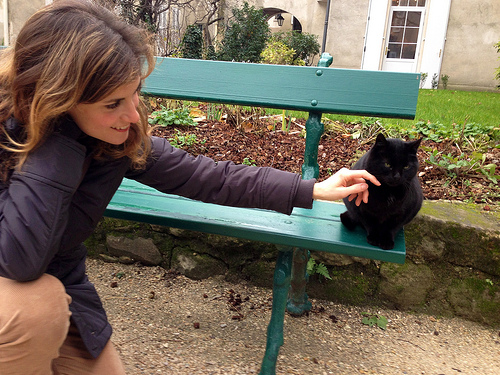}
& \includegraphics[width=0.12\textwidth,height=1.0cm]{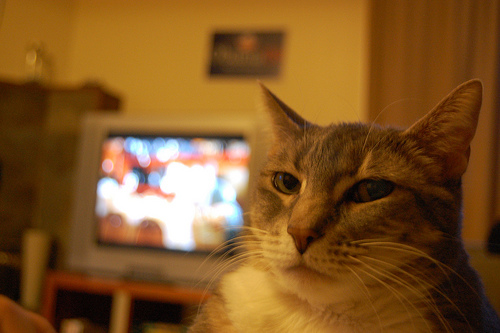} \\
\midrule
Label & 0 (dog) & 0 (dog) & 1 (cat) & 1 (cat) \\
Spurious feature & 0 (outdoor) & 1 (indoor) & 0 (outdoor) & 1 (indoor) \\
Train data & 784 (34.4\%) & 507 (22.3\%) & 196 (8.6\%) & 789 (34.7\%) \\
Val data & 127 & 75 & 33 & 114 \\
Test data & 273 & 191 & 65 & 345 \\
\bottomrule
\end{tabular}}
\label{tab:metashift_example}
\end{table}

\clearpage
\section{Experiment details}

For evaluation, we report both \emph{worst-group accuracy} and \emph{mean accuracy}. Let $\mathcal{D}_{\mathrm{test}}=\{(x_i,y_i)\}_{i=1}^{N}$ denote the evaluation set, where $x_i\in\mathcal{X}$ is the input, $y_i\in\mathcal{Y}$ is the target label, and $N$ is the total number of samples. Following Sec.~3.1, let $\mathcal{G}\subseteq\mathcal{Y}\times\mathcal{A}$ denote the set of all groups induced by the target label and spurious attribute, where each group is indexed by $g=(y,a)\in\mathcal{G}$. We denote by $\mathcal{D}_g\subseteq\mathcal{D}_{\mathrm{test}}$ the subset of samples belonging to group $g$, and by $N_g=|\mathcal{D}_g|$ its size. Given the classifier $f_{\theta}:\mathcal{X}\rightarrow\mathbb{R}^{Y}$, the accuracy on group $g$ is defined as
\begin{equation}
\mathrm{Acc}_g
=
\frac{1}{N_g}
\sum_{(x_i,y_i)\in\mathcal{D}_g}
\mathbbm{1}\!\left(\arg\max_{c\in\mathcal{Y}} f_{\theta}(x_i)_c = y_i\right).
\end{equation}
The \emph{worst-group accuracy} is then defined as
\begin{equation}
\mathrm{WorstAcc}
=
\min_{g\in\mathcal{G}} \mathrm{Acc}_g,
\end{equation}
while the \emph{mean accuracy} is defined as
\begin{equation}
\mathrm{MeanAcc}
=
\frac{1}{N}
\sum_{(x_i,y_i)\in\mathcal{D}_{\mathrm{test}}}
\mathbbm{1}\!\left(\arg\max_{c\in\mathcal{Y}} f_{\theta}(x_i)_c = y_i\right).
\end{equation}


For reproducibility, we provide the detailed training configurations in Table~\ref{tab:hyperparameter}, including the model architecture, learning rate, weight decay, batch size, number of epochs, prompt length, learning-rate scheduler, and optimizer. In all experiments, we use SGD with momentum and weight decay for optimization. To ensure fair comparison under standard experimental protocols, we use the baseline results reported in the original papers whenever available.

\begin{table}[!ht]
\centering
\scriptsize
\setlength{\tabcolsep}{4pt}
\renewcommand{\arraystretch}{1.2}
\caption{Hyperparameter settings.}
\resizebox{0.82\textwidth}{!}{%
\begin{tabular}{l l c c c c c c c}
\toprule
Dataset & Architecture & Learning Rate & Weight Decay & Batch Size & Epochs & Prompt length & Scheduler & Optimizer \\
\midrule
Waterbirds~\cite{Waterbirds} & ViT B/16 & 0.1 & 0.001 & 64 & 20 & 10 & No & SGD \\
CelebA~\cite{celeba} & ViT B/16 & 0.01 & 0.0001 & 100 & 20 & 20 & StepLR & SGD \\
CMNIST~\cite{Cmnist} & ViT B/16 & 0.1 & 0.001 & 100 & 10 & 10 & No & SGD \\
MetaShift~\cite{metashift} & ViT B/16 & 0.1 & 0.1 & 100 & 20 & 10 & StepLR & SGD \\
\bottomrule
\end{tabular}%
}
\label{tab:hyperparameter}
\end{table}

\section{Additional empirical results}

In this section, we present additional results on Waterbirds~\cite{Waterbirds}, CelebA~\cite{celeba}, MetaShift~\cite{metashift}, and CMNIST~\cite{Cmnist} to study the effect of prompt length and compare our method with representative group-unsupervised baselines, including JTT~\cite{liu2021justtraintwiceimproving}, MaskTune~\cite{asgari2022masktune}, and XRM~\cite{pezeshki2024discoveringenvironmentsxrm}. As shown in Table~\ref{tab:n_comparison}, prompt length has a clear impact on performance, and moderate prompt lengths generally yield the best trade-off between worst-group accuracy and mean accuracy. In particular, a prompt length of $10$ gives the best worst-group accuracy on Waterbirds and MetaShift, while prompt lengths of $15$ and $20$ perform best on CMNIST. On CelebA, our method also remains consistently strong across different prompt lengths. Overall, our method consistently outperforms JTT and MaskTune, and further surpasses XRM on Waterbirds and MetaShift in worst-group accuracy, while achieving competitive or better performance on CelebA and CMNIST. Results on Waterbirds, MetaShift, and CMNIST are averaged over three random seeds with standard deviations reported, whereas CelebA results are obtained from a single run. ``--'' denotes unavailable results, and ``*'' indicates results obtained from our training.


\begin{table}[!ht]
\centering
\renewcommand{\arraystretch}{1.2}
\caption{Comparison across different prompt lengths and baseline methods on four benchmarks.}
\resizebox{0.82\textwidth}{!}{%
\begin{tabular}{ll|ccccc|ccc}
\toprule
Dataset & Metric & 1 & 5 & 10 & 15 & 20 & JTT & MaskTune & XRM \\
\midrule
\multirow{2}{*}{Waterbirds}
& WGA(\%)$\uparrow$
& 84.47{\scriptsize$\pm$1.11}
& 86.71{\scriptsize$\pm$1.88}
& \textbf{90.81}{\scriptsize$\pm$0.31}
& 87.89{\scriptsize$\pm$2.27}
& 86.84{\scriptsize$\pm$2.76}
& 86.7
& 86.4*{\scriptsize$\pm$1.9}
& 88.1{\scriptsize$\pm$0.9} \\

& Mean(\%)
& 87.13{\scriptsize$\pm$1.84}
& 92.65{\scriptsize$\pm$1.17}
& 93.06{\scriptsize$\pm$0.33}
& 92.61{\scriptsize$\pm$1.54}
& \textbf{93.71}{\scriptsize$\pm$0.68}
& 93.3
& 93.0*{\scriptsize$\pm$0.7}
& 89.3 \\
\midrule

\multirow{2}{*}{CelebA}
& WGA(\%)$\uparrow$
& 91.33
& 88.33
& \textbf{92.19}
& 90.56
& 92.11
& 81.1
& 78.0{\scriptsize$\pm$1.2}
& 89.1{\scriptsize$\pm$1.3} \\

& Mean(\%)
& 92.26
& \textbf{92.72}
& 92.60
& 92.18
& 92.49
& 88.0
& 91.3{\scriptsize$\pm$0.1}
& 91.4 \\
\midrule

\multirow{2}{*}{MetaShift}
& WGA(\%)$\uparrow$
& 76.42{\scriptsize$\pm$1.85}
& 79.41{\scriptsize$\pm$3.93}
& \textbf{82.20}{\scriptsize$\pm$1.39}
& 67.02{\scriptsize$\pm$12.00}
& 75.56{\scriptsize$\pm$6.55}
& 69.6*
& 72.31*{\scriptsize$\pm$6.71}
& 77.5{\scriptsize$\pm$1.0} \\

& Mean(\%)
& 86.39{\scriptsize$\pm$0.41}
& 88.06{\scriptsize$\pm$0.43}
& \textbf{88.25}{\scriptsize$\pm$1.32}
& 81.96{\scriptsize$\pm$6.97}
& 85.85{\scriptsize$\pm$2.22}
& 78.1*
& 87.68*{\scriptsize$\pm$1.56}
& -- \\

\midrule
\multirow{2}{*}{CMNIST}
& WGA(\%)$\uparrow$
& 82.75{\scriptsize$\pm$6.26}
& 88.74{\scriptsize$\pm$0.89}
& 91.99{\scriptsize$\pm$0.98}
& \textbf{92.75}{\scriptsize$\pm$1.03}
& \textbf{92.75}{\scriptsize$\pm$2.05}
& 69.6*
& 2.17*{\scriptsize$\pm$3.77}
& 90.5*{\scriptsize$\pm$2.1} \\

& Mean(\%)
& 88.95{\scriptsize$\pm$2.51}
& 95.25{\scriptsize$\pm$1.17}
& \textbf{96.39}{\scriptsize$\pm$0.97}
& 95.72{\scriptsize$\pm$1.36}
& 95.94{\scriptsize$\pm$1.08}
& 99.6*
& 52.49*{\scriptsize$\pm$2.02}
& 98.0*{\scriptsize$\pm$0.5} \\
\bottomrule
\end{tabular}%
}
\label{tab:n_comparison}
\end{table}


\clearpage
\section{Pipeline of D2CP}

\begin{algorithm}[!ht]
\caption{Training pipeline of the proposed D2CP.}
\label{alg:d2cp}
\small
\begin{algorithmic}[1]
\State \textbf{Input:} Training set $\mathcal{D}_{\mathrm{train}}=\{(\mathbf{x}_i,y_i)\}_{i=1}^{N}$, concept number $K$, high-confidence ratio $\rho$, top-$M$, frozen backbone $\mathcal{F}_{\mathrm{V}}(\cdot)$, training epochs $T$
\State \textbf{Output:} Trained parameters $\Theta=\{\mathbf{V}^{y},\mathbf{V}^{a},\mathbf{W}^{y},\mathbf{W}^{a}\}$

\State Train EncDiff and obtain $f_C(\cdot)$ and $\Phi(\cdot)$

\Statex \texttt{// Before Training: Confidence-guided Bias Concept Mining (CBCM)}
\For{each $(\mathbf{x}_i,y_i)\in\mathcal{D}_{\mathrm{train}}$}
    \State Extract concept tokens $\mathbf{S}(\mathbf{x}_i)$ and projected concept features $\Phi(\mathbf{S}(\mathbf{x}_i))$ by Eqs.~\eqref{eq:weight_defination} and~\eqref{eq:mapping_function}
    \State Form image-level representation $\boldsymbol{\phi}(\mathbf{x}_i)=[\mathbf{E}_{i,1};\cdots;\mathbf{E}_{i,K}]$
\EndFor
\State Train biased classifier $\mathbf{A}$ on $\{(\boldsymbol{\phi}(\mathbf{x}_i),y_i)\}_{i=1}^{N}$ and compute confidence $p_A(y_i\mid \mathbf{x}_i)$
\For{each class $c\in\{0,1\}$}
    \State Select top $\rho\%$ high-confidence samples $\mathcal{I}_c^{\mathrm{hc}}$
    \State Compute class-specific prototypes $\bar{\mathbf{E}}_{c,k}$ by Eq.~\eqref{eq:bias_slot_proto}
\EndFor
\State Compute discriminative direction $\Delta\mathbf{W}$ by Eq.~\eqref{eq:delta_w}
\State Compute concept scores $r_{c,k}$ by Eq.~\eqref{eq:bias_score}
\State Select top-$M$ concepts $\mathcal{T}_c$ and obtain candidate spurious concept set $\mathcal{T}$ by Eqs.~\eqref{eq:topm_concepts} and~\eqref{eq:bias_concept_set}
\For{each $(\mathbf{x}_i,y_i)\in\mathcal{D}_{\mathrm{train}}$}
    \State Generate pseudo spurious label $\tilde{a}_i$ from concepts indexed by $\mathcal{T}$
\EndFor

\Statex \texttt{// During Training: Dual-branch Cross-projection Debiasing (DCD)}
\State Initialize prompts $\mathbf{V}^{y},\mathbf{V}^{a}$ and classifier heads $\mathbf{W}^{y},\mathbf{W}^{a}$
\For{epoch $=1$ to $T$}
    \For{mini-batch $\mathcal{B}=\{(\mathbf{x}_i,y_i,\tilde{a}_i)\}_{i=1}^{B}$}
        \State Extract target and spurious features $\mathbf{h}^{y}_i=\mathcal{F}_{\mathrm{V}}(\mathbf{x}_i,\mathbf{V}^{y})$ and $\mathbf{h}^{a}_i=\mathcal{F}_{\mathrm{V}}(\mathbf{x}_i,\mathbf{V}^{a})$
        \State Construct null-space projectors $\mathbf{P}_{0}^{y}$ and $\mathbf{P}_{0}^{a}$ from $\mathbf{W}^{y}$ and $\mathbf{W}^{a}$
        \State Cross-project features $\tilde{\mathbf{h}}^{y}_i=\mathbf{P}_{0}^{a}\mathbf{h}^{y}_i$ and $\tilde{\mathbf{h}}^{a}_i=\mathbf{P}_{0}^{y}\mathbf{h}^{a}_i$
        \State Compute sample-wise losses $\ell_i^{y}$ and $\ell_i^{a}$, and assign group index $g_i=2y_i+\tilde{a}_i$
        \For{each group $g\in\{0,1,2,3\}$}
            \State Collect group samples $\mathcal{B}_g=\{i\mid g_i=g\}$ and compute group losses $\mathcal{L}_g^{y}$ and $\mathcal{L}_g^{a}$
        \EndFor
        \State Compute GroupDRO losses $\mathcal{L}_{\mathrm{DRO}}^{y}$ and $\mathcal{L}_{\mathrm{DRO}}^{a}$
        \State Update group weights $q_g$ by Eq.~\eqref{eq:weight_update}
        \State Compute total loss $\mathcal{L}_{\mathrm{total}}=\mathcal{L}_{\mathrm{DRO}}^{y}+\mathcal{L}_{\mathrm{xrm}}^{y}+\mathcal{L}_{\mathrm{DRO}}^{a}+\mathcal{L}_{\mathrm{xrm}}^{a}$
        \State Update $\mathbf{V}^{y},\mathbf{V}^{a},\mathbf{W}^{y},\mathbf{W}^{a}$
    \EndFor
\EndFor
\State \Return $\Theta$
\end{algorithmic}
\end{algorithm}

\clearpage
\section{Additional diffusion-based qualitative results}

\begin{figure}[ht!]
    \centering
    \includegraphics[width=0.98\textwidth, height=0.9\textwidth]{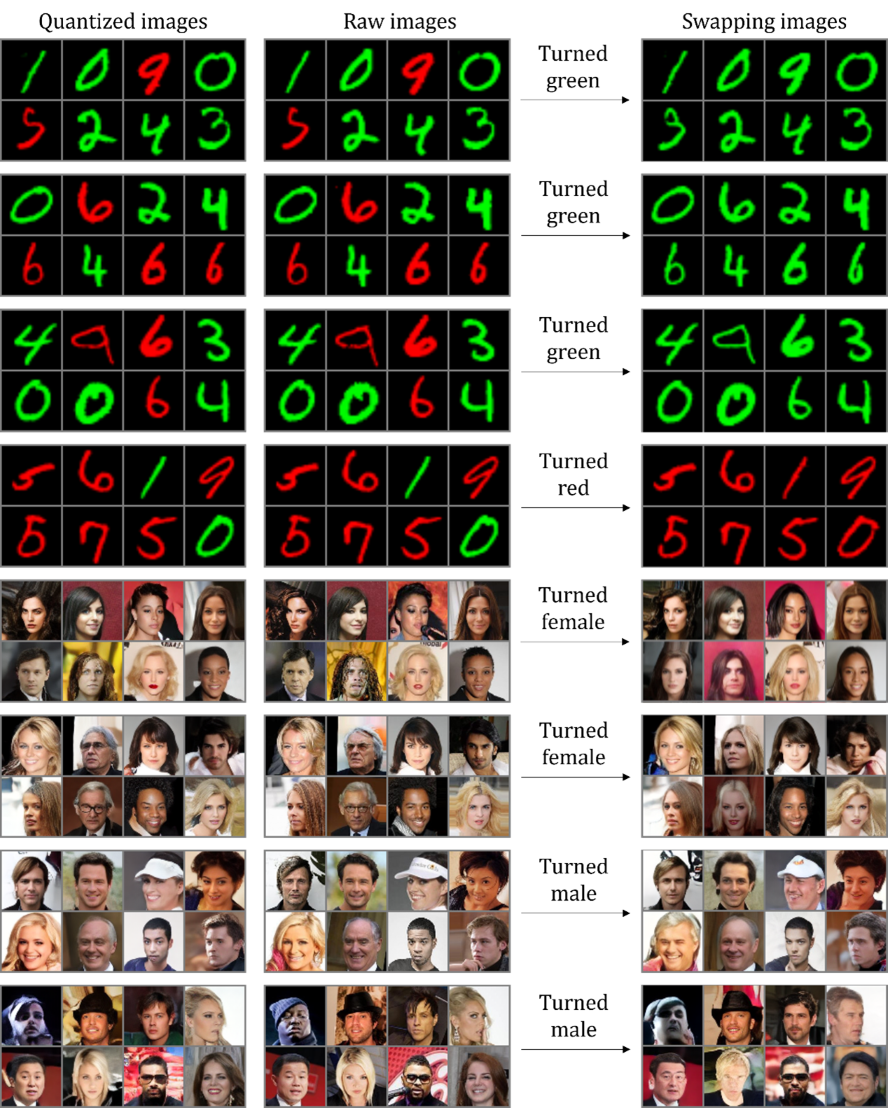}
    \caption{Additional qualitative results of concept manipulation and quantized reconstruction on CMNIST~\cite{Cmnist} and CelebA~\cite{celeba}.}

    \label{fig:diffusion_appendix}
\end{figure}

Fig.~\ref{fig:diffusion_appendix} provides additional qualitative results on CMNIST~\cite{Cmnist} and CelebA~\cite{celeba}. For each dataset, the middle column shows the original decoded samples, the right column shows results after replacing the discovered spurious concept dimension with that from another sample, and the left column shows images decoded from the corresponding quantized predicted representation. On CMNIST, the intervention consistently changes the digit color while largely preserving digit identity and stroke structure, suggesting that the discovered dimension captures a controllable spurious attribute. On CelebA, it mainly alters gender-related visual attributes while retaining most identity, pose, and facial structure. In addition, the quantized predicted representations on both datasets can still be decoded into coherent images with clear semantic content, indicating that the learned concept space remains structured after quantization. Overall, these results further support that CBCM identifies semantically meaningful and controllable spurious concepts across different data modalities.

\clearpage
\section{Additional gradCAM visualizations}

\begin{figure}[ht!]
    \centering
    \includegraphics[width=0.98\textwidth]{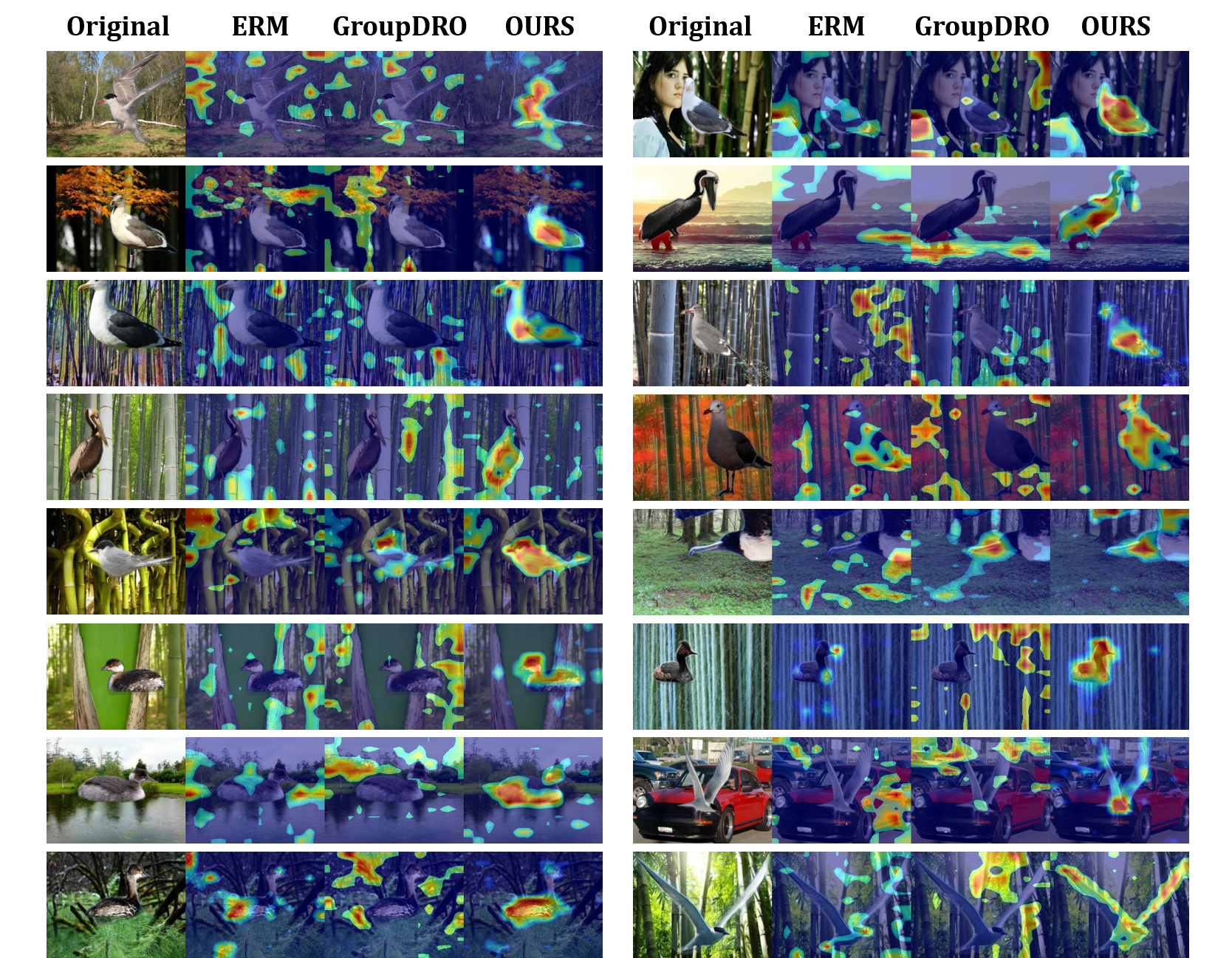}
    \caption{Waterbirds GradCAM.}

    \label{fig:heatmap}
\end{figure}

We provide additional GradCAM visualizations of ERM, GroupDRO, and D2CP on Waterbirds in Fig.~\ref{fig:heatmap}. Compared with ERM and GroupDRO, D2CP produces more object-centric attention, focusing more consistently on the bird while reducing responses to background regions. In contrast, ERM and GroupDRO often exhibit stronger activations on contextual cues such as water, vegetation, and sky. These visualizations suggest that D2CP alleviates background bias and learns representations that are better aligned with the true target object.

\end{document}